%% file: root.tex
\documentclass{article}

\usepackage[preprint]{corl_2026} 

\usepackage{graphicx}
\usepackage{booktabs}
\usepackage{amsmath}
\usepackage{amssymb}
\usepackage{array}
\usepackage{wrapfig}
\usepackage{needspace}
\usepackage{float}
\usepackage{enumitem}
\usepackage{algorithm}
\usepackage{algpseudocode}
\usepackage{pgfplots}
\usepackage[table,HTML]{xcolor}
\pgfplotsset{compat=1.18}

\definecolor{mutedyellow}{rgb}{0.922,0.800,0.455}
\definecolor{mutedolive}{rgb}{0.773,0.773,0.478}
\definecolor{mutedgreen}{rgb}{0.627,0.745,0.522}
\definecolor{mutedyellowedge}{rgb}{0.549,0.435,0.165}
\definecolor{mutedoliveedge}{rgb}{0.431,0.431,0.208}
\definecolor{mutedgreenedge}{rgb}{0.369,0.478,0.282}

\definecolor{nhred}{HTML}{D62727}

\newcommand{\method}{FACT}

\title{\Large FACT: Failure-Aware Causal Training for World-Action Models}
\newcommand{\blfootnote}[1]{%
  \begingroup
    \renewcommand{\thefootnote}{}%
    \footnote{#1}%
    \addtocounter{footnote}{-1}%
  \endgroup
}

\author{
  \normalfont
  Quanquan Peng\textsuperscript{*}, Yutong Liang\textsuperscript{*}, Rui Yan, Nicklas Hansen, Xiaolong Wang \\
  University of California San Diego \\[13pt]
  \url{https://fact-wam.github.io}
}

\begin{document}
\maketitle
\blfootnote{\textsuperscript{*} denotes equal contribution.}

\vspace{-0.4in}\vspace{-13pt}

\input{sections/abs}


\begin{figure}[H]
    \centering
    \includegraphics[width=\linewidth]{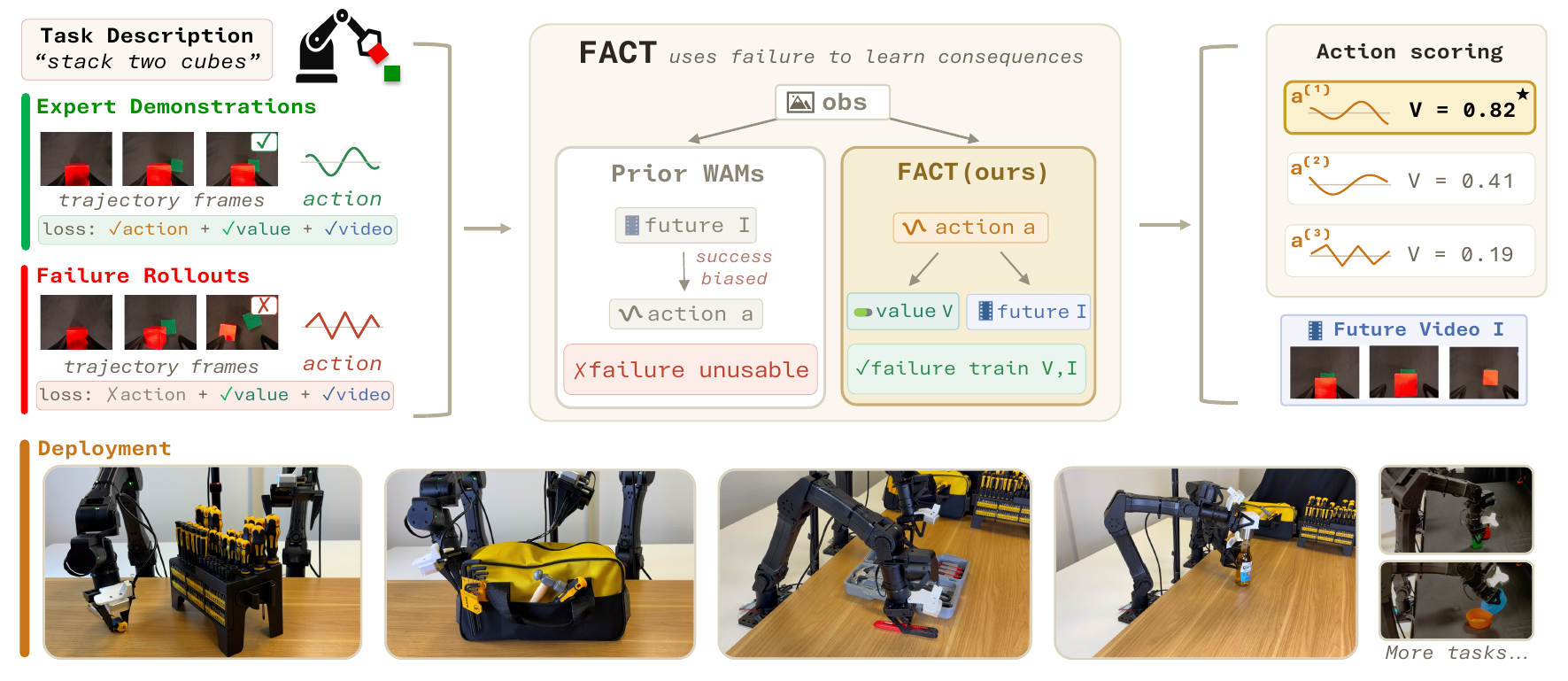}
    \vspace{-15pt}
    \caption{\textbf{\method{}} first generates an action and then rolls out the resulting future video and task-progress value. Because each future unfolds from the action that produced it, failure rollouts directly supervise the future-prediction branch under their own bad actions, teaching the world model what wrong behavior actually leads to.}
    \label{fig:teaser}
    \vspace{-10pt}
\end{figure}


\input{sections/intro}


\input{sections/related_works}


\input{sections/method}


\input{sections/exp}


\input{sections/limit}


\input{sections/conclusion}


\acknowledgments{This work was supported, in part, by NSF CAREER Award IIS-2240014 and NSF CCF-2112665 (TILOS).}


\clearpage
\bibliography{reference}  

\input{sections/appendix}

\end{document}

%% file: sections/abs.tex

\begin{abstract}
Recent world-action models (WAMs) show that co-training policies with future 
prediction can provide physical priors for action generation. 
Building on the future-prediction ability of video models, many WAMs 
generate future videos and recover actions with inverse-dynamics 
models, or use these predicted videos as goal conditions for 
action generation. 
In both cases, the world model is trained mostly on successful demonstrations 
and has little reason to predict the consequences of bad actions. 
We introduce \textbf{\method}, a causal World-Action Model 
that predicts future video and task progress conditioned on the executed action.
This action-conditioned interface allows failure rollouts to supervise action consequences, 
turning bad actions into valid future targets rather than being discarded.
Failure-aware training makes the progress predictor aware of both successful and failed action outcomes, which can optionally be used to score sampled action candidates at inference.
    Extensive experiments on simulation and real-world bimanual manipulation tasks show that \method{} outperforms many existing baselines, improves as failure data are incorporated 
    into training, and reduces success-biased future hallucination under bad actions.
\end{abstract}

%% file: sections/intro.tex

\section{Introduction}
\label{sec:intro}

Building general-purpose robot policies is a central goal in robot learning. Vision-language-action (VLA) models have emerged as a 
promising paradigm toward this aim: given image observations and a language instruction, they learn to output robot actions~\citep{brohan2023rt2,kim2024openvla,black2024pi0,physicalintelligence2025pi05}. 
A complementary line of work, world-action models (WAMs), augments this paradigm with future visual prediction, using video models or future-prediction objectives to learn control together with how the scene may evolve under robot interaction~\citep{li2026lintbot,kim2026cosmos,bi2025motus,yuan2026fast,ye2026gigaworld,hu2026bagelvla}. By coupling action modeling with predicted future observations, WAMs provide a natural way to bring temporal dynamics and physical priors into robot policy learning.

Existing WAMs commonly use predicted futures as an intermediate signal for choosing actions. 
One way is to imagine future videos and decode actions with an inverse dynamics model, as in video-first systems~\citep{li2026lintbot, ye2026world}. 
This design benefits from a strong world prior, but action decoding depends on a second-stage network and often requires the future video to be fully denoised before control can proceed. 
A second line uses predicted future frames or latents as conditions for action prediction~\citep{kim2026cosmos,yuan2026fast,hu2026bagelvla, zhou2026tau}. These designs make future prediction useful as an auxiliary signal, yet their future targets are usually expert demonstrations: the model sees plausible futures paired with good actions, 
but not the consequences of bad actions. As a result, a bad action at test time can still be paired with a success-biased future~\citep{bai2024hallucination, rawte2023survey}. The key challenge is not merely adding more data, but using rollouts that fail to complete the desired task \emph{without} turning them into bad demonstrations.

This raises the question: \textit{can we use failure rollouts as consequence supervision without treating failed actions as imitation targets?} In this work, we propose \textbf{\method} (\textbf{F}ailure-\textbf{A}ware \textbf{C}ausal \textbf{T}raining), a causal World-Action Model that separates what to imitate from what to predict. \method{} first proposes an action, then predicts the future video and task-progress value conditioned on the executed action. For successful demonstrations, actions, future video, and progress are all supervised. 
For failure rollouts, the action imitation loss is masked, but the observed failed future and lower progress value remain valid supervision. This turns failure data into action-conditioned consequence supervision and mitigates the tendency to hallucinate only successful futures under incorrect actions.

By learning action-conditioned consequences from both successful demonstrations and failed rollouts, \method{} produces a progress estimate that is sensitive to action quality. At inference time, the model can either execute the sampled action directly or optionally use this estimate to rank action candidates.
\textbf{Overall, our contributions are:}
\begin{enumerate}[itemsep=2pt, topsep=2pt, parsep=0pt, leftmargin=1.75em]
    \vspace{-0.05in}
    \item We propose \textbf{\method, a causal World-Action Model with an action-then-future sequence}, enabling co-training on failure trajectories without corrupting action decoding.
    \item We introduce a teacher-forced action-conditioned mask that \textbf{separates action generation from future prediction}, allowing failed actions to supervise future and value learning without undermining policy.
    \item \textbf{We validate the model in simulation and real-world benchmarks}, showing improved policy success, reduced success-biased future hallucination under bad actions, and a progress predictor that can optionally support candidate ranking.
\end{enumerate}

%% file: sections/related_works.tex

\section{Related Work}
\label{sec:related}

\paragraph{Vision-language-action models for robotic control.}
Vision-language-action (VLA) models have emerged as a leading paradigm for generalist robot policies, transferring semantic priors from vision-language models to low-level action generation. Early large-scale robot transformers and open-source generalist policies learn language-conditioned manipulation from heterogeneous robot datasets~\citep{brohan2023rt2,kim2024openvla,brohan2022rt,team2024octo}, while recent flow- or diffusion-based policies further scale continuous action modeling and open-world generalization~\citep{black2024pi0,physicalintelligence2025pi05,zheng2025xvla,wu2026pragmatic,lingbotvla2}. 
Beyond direct action prediction, a growing line of work incorporates predictive dynamics into VLA policies. BagelVLA~\citep{hu2026bagelvla} interleaves language reasoning, visual forecasting, and action generation; 
DreamZero~\citep{ye2026world} and other world-action models use future visual prediction for action decoding, planning, or data generation~\citep{li2026lintbot,kim2026cosmos,bi2025motus,yuan2026fast,ye2026gigaworld,gao2026dreamdojo,lyu2026lda,jiang2026cross,liu2026long}. These methods show that robotic policies can benefit from modeling not only what action to execute, but also how the world may evolve. In contrast, our work focuses on the causal role of action-conditioned futures: failure rollouts provide structured supervision about undesirable consequences rather than merely serving as low-quality demonstrations.

\paragraph{Data sources for robotic policy learning.}
The progress of robotic foundation models is closely tied to broader and more diverse training data. One line of work scales real robot demonstrations through shared datasets, distributed collection, and low-cost teleoperation platforms~\citep{dasari2019robonet,walke2023bridgedata,o2024open,khazatsky2024droid,zhao2023learning,fu2024mobile}. 
Another line reduces collection cost by using human-centric interfaces or egocentric human videos to transfer manipulation priors to robots~\citep{chi2024universal,xu2025dexumi,cheng2026tacumi,yang2025egovla,zheng2026egoscale,qiu2025humanoid}. Simulation provides complementary supervision through scalable task generation and language-conditioned benchmarks~\citep{nasiriany2024robocasa,nasiriany2026robocasa365,mandlekar2023mimicgen,liu2023libero,mees2022calvin,chen2025robotwin2}. Recently, reinforcement learning and self-improvement have also re-emerged as alternatives to purely offline imitation, improving policies through rollouts, verified rewards, value models, or world-model-based interaction~\citep{li2025vla,lv2026viva,yang2026rise}. \method{} follows this broader trend of extracting supervision beyond successful demonstrations, but specifically studies how failure data~\citep{liu2023reflect,grollman2012robot,wang2026learning,li2026recover} can be converted to action-conditioned future and value supervision for robust policy learning.

%% file: sections/method.tex

\section{Method}
\label{sec:method}

We propose \method{}, a causal World-Action Model that generates actions before predicting their consequences. 
In this section, we first define the task setting and prediction targets (Sec.~\ref{sec:method_problem}), and
then introduce an action-conditioned architecture that separates action imitation from future prediction, 
allowing failure rollouts to supervise the world branch without becoming imitation targets (Sec.~\ref{sec:method_arch}). 
Finally, the training and inference strategy closes this loop: failure rollouts provide supervision about action 
consequences, and the learned progress estimate provides an optional interface for scoring candidate actions 
(Sec.~\ref{sec:method_training}).

\subsection{Problem Formulation}
\label{sec:method_problem}

We formulate language-conditioned robotic manipulation as a sequential decision-making problem. At time $t$, the robot receives a task instruction $\ell$ and the current observation
$o_t = (I^{\mathrm{main}}_t, I^{\mathrm{wristL}}_t, I^{\mathrm{wristR}}_t, s_t),$
where the images are multi-view RGB observations and $s_t \in \mathbb{R}^{d_s}$ is the robot proprioceptive state. The goal is to choose an action chunk $a_{t:t+H} \in \mathbb{R}^{H \times d}$ that advances the task specified by $\ell$. A standard language-conditioned policy models this decision directly as
\begin{equation}
    p_\theta(a_{t:t+H} \mid o_t, \ell).
    \label{eq:action_policy}
\end{equation}
Meanwhile, the model is additionally asked to predict what the chosen action will lead to. We therefore augment Eq.~\eqref{eq:action_policy} with an action-conditioned world branch:
\begin{equation}
    p_\theta(o'_{t:t+K}, v_t(a_{t:t+H}) \mid o_t, \ell, a_{t:t+H}),
    \label{eq:world_policy}
\end{equation}
where $o'_{t:t+K}$ denotes future-video observations over the prediction horizon and $v_t(a_{t:t+H}) \in [0,1]$ is the predicted task progress after executing action chunk $a_{t:t+H}$. 
To keep this value target comparable across episodes with different lengths, we first define a normalized progress variable from an episode-level return. Let $r_k$ be a progress reward and $G_t=\sum_{k=1}^{t} r_k$ be the cumulative progress up to time $t$. The normalized progress is
\begin{equation}
    p_t = \frac{G_t}{G_T} \in [0,1],
    \label{eq:value_general}
\end{equation}
where $p_t=0$ denotes the beginning of the task and $p_t=1$ denotes completion. The learned value head is queried as $V_\theta(o_t,\ell,a_{t:t+H})$ and predicts the action-conditioned progress target.

\subsection{Model Architecture}
\label{sec:method_arch}

\paragraph{Action-conditioned future prediction.}
The model must learn future outcomes from failure rollouts without imitating failed actions. To separate these two effects, we introduce a clean ground-truth action condition $a^{\mathrm{gt}}_{t:t+H}$ for the world branch. Under this teacher-forcing design, Eq.~\eqref{eq:world_policy} becomes
\begin{equation}
    p_\theta(o'_{t:t+K}, v_t(a^{\mathrm{gt}}_{t:t+H}) \mid o_t, \ell, a^{\mathrm{gt}}_{t:t+H}).
    \label{eq:teacher_forced_world}
\end{equation}

\begin{figure}[t]
    \centering
    \includegraphics[width=\linewidth]{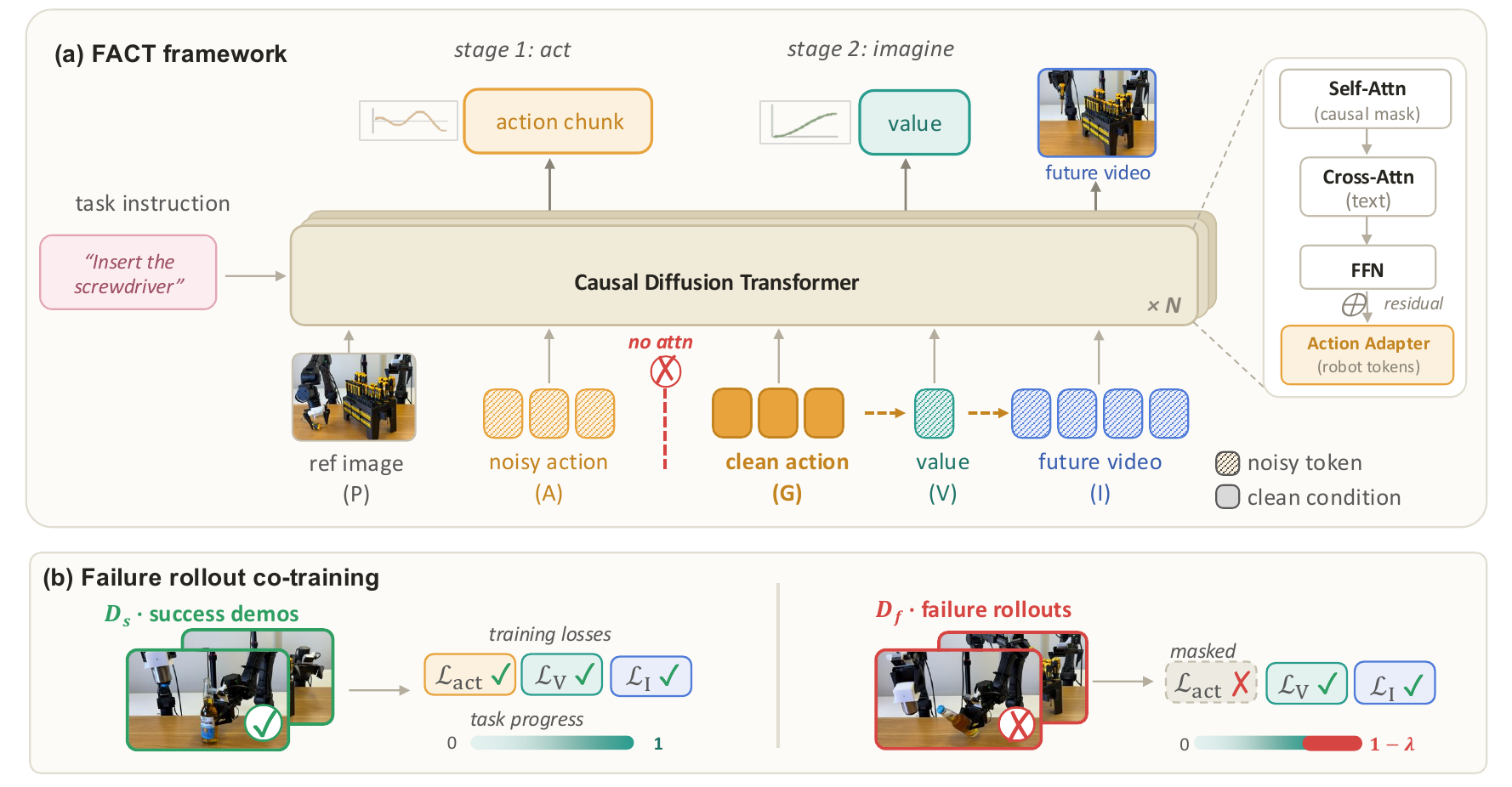}
    \vspace{-0.15in}
    \caption{\textbf{\method{} architecture.}
    \textbf{(a)} A shared causal diffusion transformer denoises action, value, and future-video tokens; value and future video condition on the clean action slot $G$, not the noisy $A$. At inference, Stage~1 denoises an action chunk, which fills $G$ for Stage~2 value and future-video prediction.
    \textbf{(b)} Success demonstrations supervise all three losses; failure rollouts mask $\mathcal{L}_{\mathrm{act}}$ but keep value and future-video supervision with a lowered progress target---failures teach consequences, not behavior.}
    \label{fig:arch}
    \vspace{-10pt}
\end{figure}

This teacher-forced action token conditions future-video and value prediction, while a separate predicted-action segment is used for action denoising. \method{} implements this factorization with the token order
\begin{equation}
z = [z^P_{\mathrm{ref}} \,\|\, z^A_{\mathrm{pred}} \,\|\, z^G_{\mathrm{gt}} \,\|\, z^V_{\mathrm{value}} \,\|\, z^I_{\mathrm{future}}],
\label{eq:sequence}
\end{equation}

\begin{wrapfigure}[14]{r}{0.60\linewidth}
    \vspace{-10pt}
    \centering
    \includegraphics[width=0.9\linewidth]{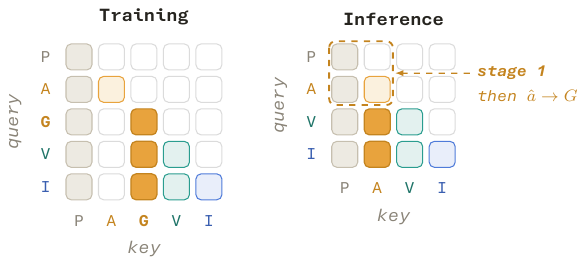}
    \vspace{-10pt}
    \caption{\textbf{Training and inference masks.} Rows and columns denote prefix image ($P$), predicted action ($A$), ground-truth action ($G$), value ($V$), and future video ($I$); cells mark allowed attention, colored by the attended segment. The dashed block is Stage~1 action denoising; the denoised action then fills $G$ for Stage~2 (Sec.~\ref{sec:method_inference}).}
    \label{fig:arch_masks}
    \vspace{-10pt}
\end{wrapfigure}
where $P$ is the observation prefix, $A$ is the noisy predicted-action segment, $G$ is the clean teacher-forced action segment, $V$ is the value segment, and $I$ is the future-video segment. World-side predictions depend on $G$ instead of $A$: successful demonstrations supervise both action and world prediction, while failure trajectories mask the action imitation loss but keep future-video and value supervision active.

The attention mask in Figure~\ref{fig:arch_masks} enforces this separation. During training, value and future-video tokens attend to the clean ground-truth action $G$ rather than the noisy predicted action $A$, while $A$ cannot see $G$. During inference, no $G$ segment is available: Stage 1 denoises $A$ from $[P,A]$, and Stage 2 denoises value and future-video tokens conditioned on the clean action from Stage 1.

\paragraph{Action adapter on a shared video backbone.}
\method{} uses a shared video diffusion transformer for action, value, and future-video prediction (Figure~\ref{fig:arch}). We attach a lightweight action adapter to robot tokens after the feed-forward network in each transformer block, giving the model action-specific capacity while preserving the pretrained video pathway. 
Since failure rollouts supervise only future video and value, this shared backbone lets future-outcome losses affect action generation instead of being confined to a separate world expert, as in Mixture-of-Transformers (MoT) designs~\citep{li2026lintbot,bi2025motus,hu2026bagelvla}.

\subsection{Training and Inference Strategy}
\label{sec:method_training}

\paragraph{Failure-aware value targets.}
We denote successful demonstrations by $\mathcal{D}_s$ and failure rollouts by $\mathcal{D}_f$. Each episode is annotated with its final outcome and, for failed episodes, the failure onset. We instantiate Eq.~\eqref{eq:value_general} as an action-conditioned progress target:
\begin{equation}
    v_t(a^{\mathrm{gt}}_{t:t+H}) =
    \begin{cases}
        p_{t+H}, & \text{if success},\\
        \mathrm{clip}\!\left(p_{t+H}-\lambda_{\mathrm{fail}}\mathbf{1}_{\mathrm{fail}}(t+H),0,1\right), & \text{if fail},
    \end{cases}
    \label{eq:failure_value}
\end{equation}
where $\mathbf{1}_{\mathrm{fail}}(t+H)$ indicates whether failure happened by executing $a^{\mathrm{gt}}_{t:t+H}$. In our experiments, we use a uniform progress reward for simplicity, so Eq.~\eqref{eq:value_general} reduces to $p_t=t/T$ for an episode of length $T$. This target preserves temporal progress while lowering the score assigned to action-conditioned futures that enter failure. Thus, under a failed action sequence, the model is trained to predict both the observed failed future and a lower action-conditioned progress target.

\paragraph{Joint denoising losses.}
We train all predicted modalities with the same flow-matching denoising objective~\citep{lipman2022flow}. For a target modality $x\in\{a,v,I\}$, let $z_0^x$ be the clean target token and $z_1^x\sim\mathcal{N}(0,I)$ be Gaussian noise. With interpolation $z_\tau^x=(1-\tau)z_0^x+\tau z_1^x$, the modality loss is
\begin{equation}
    \mathcal{L}_x =
    \mathbb{E}_{z_0^x,z_1^x,\tau}
    \left[
    \left\lVert u_\theta^x(z_\tau^x,\tau;z) - (z_1^x-z_0^x)\right\rVert_2^2
    \right],
\end{equation}
where $u_\theta^x$ is the predicted velocity and $z$ denotes the full token sequence in Eq.~\eqref{eq:sequence}. Success demonstrations supervise action, value, and future prediction:
\begin{equation}
    \mathcal{L}_{\mathcal{D}_s} = w_a\mathcal{L}_a + w_v\mathcal{L}_v + w_I\mathcal{L}_I.
\end{equation}
For failure rollouts, we remove the action imitation term while retaining value and future-video supervision:
\begin{equation}
    \mathcal{L}_{\mathcal{D}_f} = w_v\mathcal{L}_v + w_I\mathcal{L}_I.
\end{equation}
Thus, failure data teaches the model the consequences and lower action-conditioned progress of failed actions without making those actions policy targets.

\paragraph{Two-stage inference.}
\label{sec:method_inference}
Inference runs the same transformer in two stages. 
Stage 1 denoises $[P_{\mathrm{state}},P_{\mathrm{ref}},A_{\mathrm{noisy}}]$ 
for $K_{\mathrm{denoise}}$ flow-Euler steps and returns a clean action 
chunk $\hat{a}_{t:t+H}$ (in Figure~\ref{fig:arch_masks}). If only an action is required, \method{} stops here and skips world prediction. When candidate scoring or consequence prediction is requested, Stage 2 places $\hat{a}_{t:t+H}$ into the clean action-conditioning slot and denoises value and, optionally, future-video latents. Since the prefix $P$ is shared across candidates, prefix key-value caching is used in action-only inference.

\paragraph{Optional candidate scoring.}
Candidate ranking~\citep{qi2026inference} is meaningful in \method{} because the progress predictor is trained on action-conditioned failed outcomes. With success-only training, progress prediction is calibrated mainly on the expert action manifold and may assign overly optimistic scores to poor actions. We therefore treat value-guided selection as an optional deployment interface and diagnostic of failure-aware consequence learning, rather than an independent source of supervision. 
For value-guided inference, Stage 1 samples $N$ action candidates $\{a^{(k)}\}_{k=1}^{N}$ in parallel. Stage 2 predicts an action-conditioned progress score $\hat{v}^{(k)}=V_\theta(o_t,\ell,a^{(k)})$ for each candidate, and the executed action is selected as
\begin{equation}
    a^\star = \arg\max_{k \in \{1,\ldots,N\}} \hat{v}^{(k)}
    = \arg\max_{a \in a^{(1:N)}} V_\theta(o_t, \ell, a).
\end{equation}
This selection rule uses the value head to score the future implied by each candidate action, without training a separate critic.

\paragraph{Implementation details.}
We initialize the model weights from WAN2.2-5B~\citep{wan2025wan}, which serves as the video diffusion backbone, and train with AdamW~\citep{loshchilov2017decoupled}. 
The learning rates are $2\times10^{-4}$ for the action FFN and $2\times10^{-5}$ for the WAN backbone; loss weights are $w_a=20$ and $w_v=w_I=1$. 
The action chunk length is $H=48$ and $\lambda_{\text{fail}}=1$. Future-video supervision uses the current frame plus four future offsets, corresponding to $[0,H/4,H/2,3H/4,H]$. Unless otherwise specified, inference uses 20 flow-Euler denoising steps.
Appendix~\ref{app:algorithms} summarizes the failure-aware co-training loop, rollout failure collection, and two-stage inference procedure in pseudocode.

%% file: sections/exp.tex

\section{Experiments}
\label{sec:exp}
 
In this section, we evaluate \method{} through benchmark comparisons, controlled variants, and diagnostic analyses.
We first describe the experimental setup and compare our model with robot foundation policies and recent WAMs in simulation (Sec.~\ref{sec:exp_sim}) and real-world benchmarks (Sec.~\ref{sec:exp_real}).
We further analyze the model variants in Sec.~\ref{sec:exp_ablation}, focusing on how each design choice affects the overall results.
Our experiments are organized around three key questions:
\begin{enumerate}[itemsep=2pt, topsep=2pt, parsep=0pt, leftmargin=1.75em]
    \vspace{-0.05in}
    \item How does \method{} compare with existing WAMs and robot foundation policies across simulation and real-world settings?
    \item Does action-conditioned future prediction benefit action generation?
    \item Does incorporating failure data improve training and reduce the tendency to hallucinate success-biased futures?
\end{enumerate}

\paragraph{Experimental Setup}
\label{sec:exp_setup}
We evaluate \method{} in both simulation and real-world bimanual manipulation. Across settings, failure data are mainly collected from model rollouts and used as additional consequence supervision. We compare against representative robot foundation policies and recent WAM baselines.
Additional real-world task details are provided in Appendices~\ref{app:task_overview} and~\ref{app:real_setup}.

\subsection{Simulation Results on RoboTwin}
\label{sec:exp_sim}

\begin{wraptable}{r}{0.53\linewidth}
\vspace{-50pt}
\centering
\setlength{\tabcolsep}{2pt}
\scriptsize
\caption{\textbf{RoboTwin} simulation results.}
\vspace{5pt}
\label{tab:sim_results}
\resizebox{0.86\linewidth}{!}{%
\begin{tabular}{@{}lccc@{}}
\toprule
\textbf{Method} & \textbf{Clean} & \textbf{Rand.} & \textbf{Average} \\
\midrule
\multicolumn{4}{l}{\textit{External baselines}} \\
\cmidrule(lr){1-4}
$\pi_0$~\citep{black2024pi0}             & 65.9 & 58.4 & 62.2 \\
X-VLA~\citep{zheng2025xvla}              & 72.9 & 72.8 & 72.9 \\
$\pi_{0.5}$~\citep{physicalintelligence2025pi05} & 82.7 & 76.8 & 79.8 \\
Gigaworld-Policy~\citep{ye2026gigaworld} & 87.0 & 85.0 & 86.0 \\
Motus~\citep{bi2025motus}                & \textbf{88.7} & \textbf{87.0} & \textbf{87.8} \\
\midrule
\multicolumn{4}{l}{\textit{\method{} variants}} \\
\cmidrule(lr){1-4}
Ours                                      & 86.3 & 84.9 & 85.6 \\
Ours w/ failure                          & \underline{88.4} & \underline{86.6} & \underline{87.5} \\
\midrule
\multicolumn{4}{l}{\textit{Ablations}} \\
\cmidrule(lr){1-4}
Ours w/o video co-train                  & 82.5 & 81.0 & 81.8 \\
\bottomrule
\end{tabular}
}
\vspace{-30pt}
\end{wraptable}

In simulation, we train on 50 RoboTwin tasks and add about 1.3K rollout failures. Table~\ref{tab:sim_results} shows that video co-training improves
\method{} from $81.8\%$ to $85.6\%$ average success, and failure co-training further improves it to $87.5\%$.
This brings \method{} close to Motus on this benchmark ($87.5\%$ vs. $87.8\%$), while running roughly $3\times$ faster at deployment (see Appendix~\ref{app:inference_time}).
Appendix~\ref{app:robotwin} reports the per-task RoboTwin results over all 50 tasks.

\subsection{Real-World Results}
\label{sec:exp_real}

\begin{figure}[t]
    \centering
    \vspace{-10pt}
    \includegraphics[width=\linewidth]{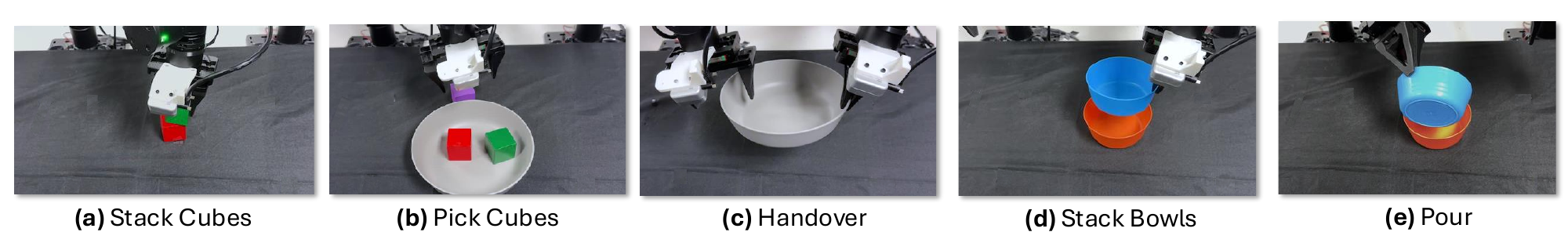}
    \vspace{-15pt}
    \caption{\textbf{Real-world task overview.} We show intermediate frame results for five tasks.}
    \label{fig:real_task_overview}
    \vspace{-8pt}
\end{figure}

We evaluate five seen tasks (Fig.~\ref{fig:real_task_overview}) and three held-out unseen variants that change object colors, shapes, and corresponding instructions. Cube-manipulation tasks use 200 expert demonstrations, the remaining seen tasks use 50, and we collect $\sim$30 failure rollouts per cube task for co-training. Each cell in Tables~\ref{tab:real_robot_seen} and~\ref{tab:real_robot_unseen} averages 20 trials; rows marked \textit{optional} use $N{=}4$ candidate scoring, while other variants use single-sample inference.

\begin{table}[H]
\centering
\begin{minipage}[t]{0.57\linewidth}
    \centering
    \setlength{\tabcolsep}{1.5pt}
    \renewcommand{\arraystretch}{0.9}
    \scriptsize
    \caption{Real-world results on \textbf{seen} tasks.}
    \label{tab:real_robot_seen}
    \resizebox{\linewidth}{!}{%
    \begin{tabular}{lccccc|c}
    \toprule
    \textbf{Method}
    & \textbf{\shortstack{Stack\\Cubes}}
    & \textbf{\shortstack{Pick\\Cubes}}
    & \textbf{\shortstack{Hand\\over}}
    & \textbf{\shortstack{Stack\\Bowls}}
    & \textbf{Pour}
    & \textbf{Avg.} \\
    \midrule
    \multicolumn{7}{l}{\textit{External baselines}} \\
    \cmidrule(lr){1-7}
    Cosmos~\citep{kim2026cosmos}        &  5 & 45 & 25 & 35 & 15 & 25 \\
    $\pi_0$~\citep{black2024pi0}        & 35 & 70 & 40 & 50 & 45 & 48 \\
    $\pi_{0.5}$~\citep{physicalintelligence2025pi05} & \underline{75} & \textbf{100} & 85 & 80 & \textbf{100} & 88 \\
    Motus~\citep{bi2025motus}           & 50 & 70 & 55 & 85 & 60 & 64 \\
    \midrule
    \multicolumn{7}{l}{\textit{\method{} variants}} \\
    \cmidrule(lr){1-7}
    Ours                                & 70 & 85 & \underline{90} & 80 & 85 & 82 \\
    Ours w/ failure                     & 75 & \underline{95} & 85 & \underline{95} & \underline{95} & \underline{89} \\
    \shortstack[l]{Ours w/ failure\\+ scoring (optional)} & \textbf{85} & \textbf{100} & 85 & \textbf{100} & 90 & \textbf{92} \\
    \midrule
    \multicolumn{7}{l}{\textit{Ablations}} \\
    \cmidrule(lr){1-7}
    \shortstack[l]{Ours + scoring} & \underline{80} & 80 & 70 & 90 & 75 & 79 \\
    Ours w/o causal mask                & 50 & 75 & \textbf{95} & 85 & 80 & 77 \\
    \shortstack[l]{Ours w/ failed-action loss} & 45 & 55 & 75 & 65 & 75 & 63 \\
    Ours w/o video co-train             & 60 & 55 & 35 & 85 & 55 & 58 \\
    \bottomrule
    \end{tabular}
    }
\end{minipage}
\hfill
\begin{minipage}[t]{0.42\linewidth}
    \centering
    \setlength{\tabcolsep}{2pt}
    \renewcommand{\arraystretch}{0.9}
    \scriptsize
    \caption{\textbf{Unseen} real-world task results.}
    \label{tab:real_robot_unseen}
    \resizebox{\linewidth}{!}{%
    \begin{tabular}{lccc|c}
    \toprule
    \textbf{Method}
    & \textbf{\shortstack{Stack\\Cubes}}
    & \textbf{\shortstack{Pick\\Cubes}}
    & \textbf{\shortstack{Stack\\Bowls}}
    & \textbf{Avg.} \\
    \midrule
    \multicolumn{5}{l}{\textit{External baselines}} \\
    \cmidrule(lr){1-5}
    Cosmos~\citep{kim2026cosmos}        & 15 & 10 &   0 &  8 \\
    $\pi_0$~\citep{black2024pi0}        & 30 & 65 &  75 & 57 \\
    $\pi_{0.5}$~\citep{physicalintelligence2025pi05} & \textbf{65} & \underline{90} & \textbf{100} & \textbf{85} \\
    Motus~\citep{bi2025motus}           & 55 & 60 &  70 & 62 \\
    \midrule
    \multicolumn{5}{l}{\textit{\method{} variants}} \\
    \cmidrule(lr){1-5}
    Ours                                  & 45 & 75 & 80 & 67 \\
    Ours w/ failure                       & \underline{60} & 85 & \underline{85} & 77 \\
    \shortstack[l]{Ours w/ failure\\+ scoring (optional)} & \textbf{65} & \textbf{95} & \underline{85} & \underline{82} \\
    \bottomrule
    \end{tabular}
    }
\end{minipage}
\vspace{-12pt}
\end{table}

On seen tasks, \method{} outperforms Motus ($82\%$ vs. $64\%$); failure-aware training raises this from $82\%$ to $89\%$, and optional scoring further reaches $92\%$. Notably, scoring alone without failed outcomes does not help ($79\%$), confirming that the value head only becomes useful after consequence training. On unseen variants, failure-aware training raises success from $67\%$ to $77\%$ and optional scoring to $82\%$, close to $\pi_{0.5}$ at $85\%$ despite $\pi_{0.5}$ benefiting from large-scale robot pretraining that \method{} does not use. Appendix~\ref{app:inference_time} compares these with measured inference latency.

\subsection{Ablation Studies}
\label{sec:exp_ablation}

\paragraph{Policy-performance ablations.}
The controlled variants in Tables~\ref{tab:sim_results} and~\ref{tab:real_robot_seen} isolate two design choices of \method{}. Removing video co-training reduces RoboTwin average success from $85.6\%$ to $81.8\%$ and real-world seen-task success from $82\%$ to $58\%$, indicating that future prediction acts as an important regularizer for action generation.
The causal-mask ablation is trained without failure data, matching the `Ours' setting except that it removes the clean ground-truth action condition $G$ in Figure~\ref{fig:arch_masks} and jointly denoises action, value, and future-video tokens. Its lower real-world seen-task success, from $82\%$ to $77\%$, suggests that teacher-forced action conditioning is important for turning future prediction into stronger action generation. When failure rollouts are added but their action imitation loss is not masked, success drops to $63\%$, confirming that failed actions should supervise consequences rather than action generation.

\paragraph{Failure data reduces future hallucination.}
\begin{wraptable}{r}{0.48\linewidth}
\vspace{-10pt}
\centering
\setlength{\tabcolsep}{4pt}
\scriptsize
\caption{\textbf{Future prediction quality (PSNR $\uparrow$).} Future-image prediction on 512 held-out samples, split evenly between success demonstrations and failure rollouts.}
\label{tab:future_prediction_metrics}
\vspace{6pt}
\scalebox{1.2}[1.1]{%
\begin{tabular}{@{}lcc@{}}
\toprule
& \textbf{Ours} & \textbf{Ours w/ failure} \\
\midrule
All & 22.82 & \textbf{26.00} \\
Success-rollout & \textbf{26.12} & 26.08 \\
Failure-rollout & \cellcolor{red!12}19.51 & \textbf{25.92} \\
\bottomrule
\end{tabular}
}
\end{wraptable}

\begin{figure}[H]
    \centering
    \vspace{-10pt}
    \includegraphics[width=0.9\linewidth]{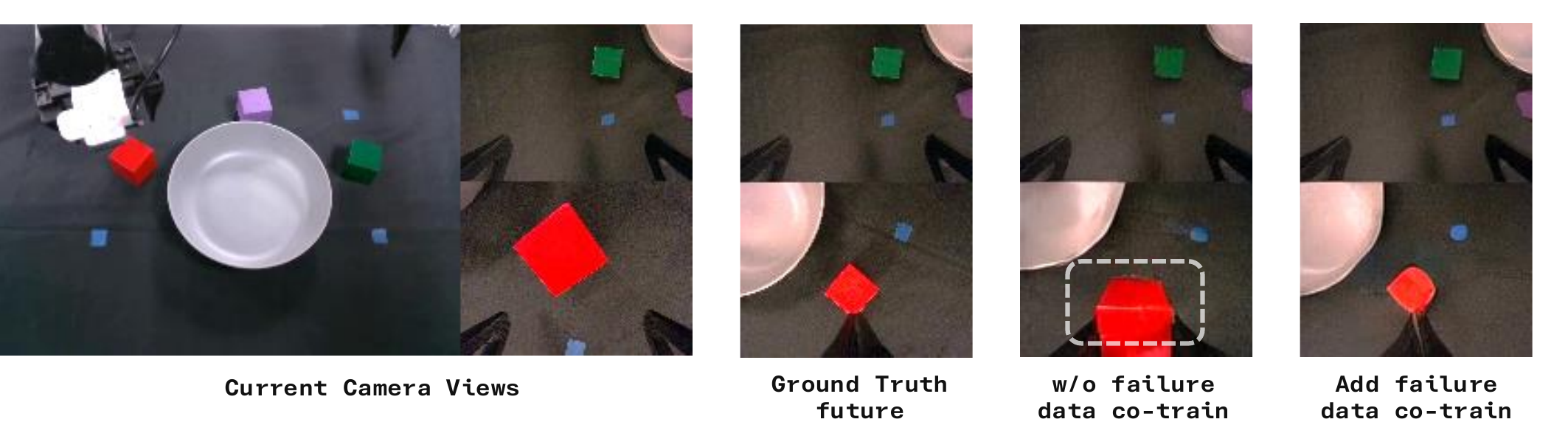}
    \vspace{-10pt}
    \caption{\textbf{Effect of failure data on future prediction.} Under the same bad-action condition, the success-only model hallucinates a successful grasp (marked by a white dotted box), while failure-aware co-training predicts the observed failure outcome.}
    \label{fig:future_pred}
\end{figure}

We compare a success-only checkpoint with one co-trained on success demonstrations and real failure rollouts. Figure~\ref{fig:future_pred} shows that, under the same bad-action condition, the success-only model still predicts a successful grasp, while failure-aware co-training predicts the observed failed outcome.
Table~\ref{tab:future_prediction_metrics} quantifies this effect: failure-aware co-training substantially improves prediction quality on failure-rollout futures while leaving successful-demonstration futures nearly unchanged, indicating that failure data reduces success-biased future hallucination without degrading normal future prediction. The same trend holds under SSIM, which we report in Appendix~\ref{app:future_metrics}.

\paragraph{Failure-data scaling.}
\begin{wrapfigure}[10]{r}{0.475\linewidth}
\vspace{-20pt}
\centering
\begin{tikzpicture}
\begin{axis}[
    width=\linewidth,
    height=3.7cm,
    ybar=1pt,
    bar width=4pt,
    enlarge x limits=0.22,
    ymin=0, ymax=105,
    ylabel={Success rate (\%)},
    ylabel style={font=\scriptsize, yshift=-4pt},
    symbolic x coords={Handover, Microwave, Bottles, Avg.},
    xtick=data,
    xticklabels={{Handover\\Block}, {Open\\Microwave}, {Put Bottles\\Dustbin}, {Avg.}},
    xticklabel style={font=\scriptsize, align=center, yshift=-2pt},
    yticklabel style={font=\scriptsize},
    ytick={0,25,50,75,100},
    legend style={
        at={(0.5,1.02)}, anchor=south,
        legend columns=3,
        font=\scriptsize,
        draw=none, fill=none,
        column sep=4pt,
    },
    legend image code/.code={\draw[#1] (0cm,-0.05cm) rectangle (0.18cm,0.12cm);},
    legend cell align=left,
    tick align=outside,
    axis line style={-,line width=.4pt},
    major grid style={line width=.2pt,draw=gray!20},
    ymajorgrids=true,
]
\addplot+[fill=mutedyellow, draw=mutedyellowedge] coordinates {
    (Handover, 14) (Microwave, 62) (Bottles, 22) (Avg., 32.7)};
\addplot+[fill=mutedolive, draw=mutedoliveedge] coordinates {
    (Handover, 38) (Microwave, 80) (Bottles, 16) (Avg., 44.7)};
\addplot+[fill=mutedgreen, draw=mutedgreenedge] coordinates {
    (Handover, 46) (Microwave, 90) (Bottles, 36) (Avg., 57.3)};
\legend{$p{=}0\%$, $p{=}50\%$, $p{=}100\%$}
\end{axis}
\end{tikzpicture}
\vspace{-8pt}
\caption{\textbf{Failure-data scaling on RoboTwin.} Success rate vs.\ failure-rollout fraction $p$.}
\label{fig:failure_scaling}
\vspace{-12pt}
\end{wrapfigure}
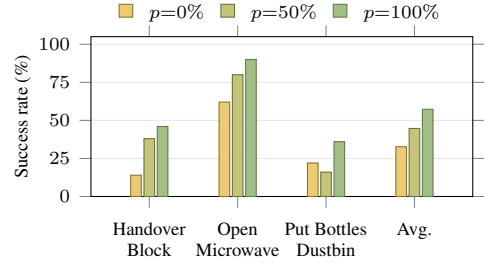
We further probe how the amount of failure data shapes policy learning. On three RoboTwin \textit{clean} tasks, we train \method{} with $p\in\{0\%,50\%,100\%\}$ of the available failure rollouts mixed into training; at $p{=}100\%$, failure data accounts for about $45\%$ of the total training set. As shown in Figure~\ref{fig:failure_scaling}, average success improves monotonically from $32.7\%$ to $57.3\%$, indicating that \method{} continues to benefit as more failure rollouts are added rather than saturating early.

\begin{figure}[b]
    \centering
    \begin{minipage}[t]{0.47\linewidth}
        \centering
        \includegraphics[ width=\linewidth]{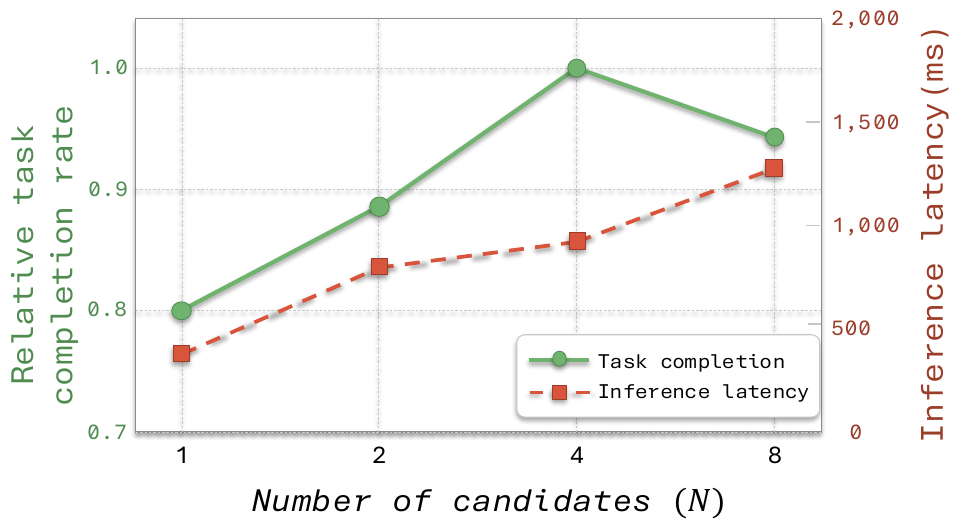}
        \vspace{-12pt}
        \caption{\textbf{Choices of different $N$.} Relative task completion and inference latency as the candidate count $N$ varies on a long-horizon grasping task; $N{=}4$ balances the two.}
        \label{fig:best_of_n}
    \end{minipage}
    \hfill
    \begin{minipage}[t]{0.47\linewidth}
        \centering
        \includegraphics[width=\linewidth]{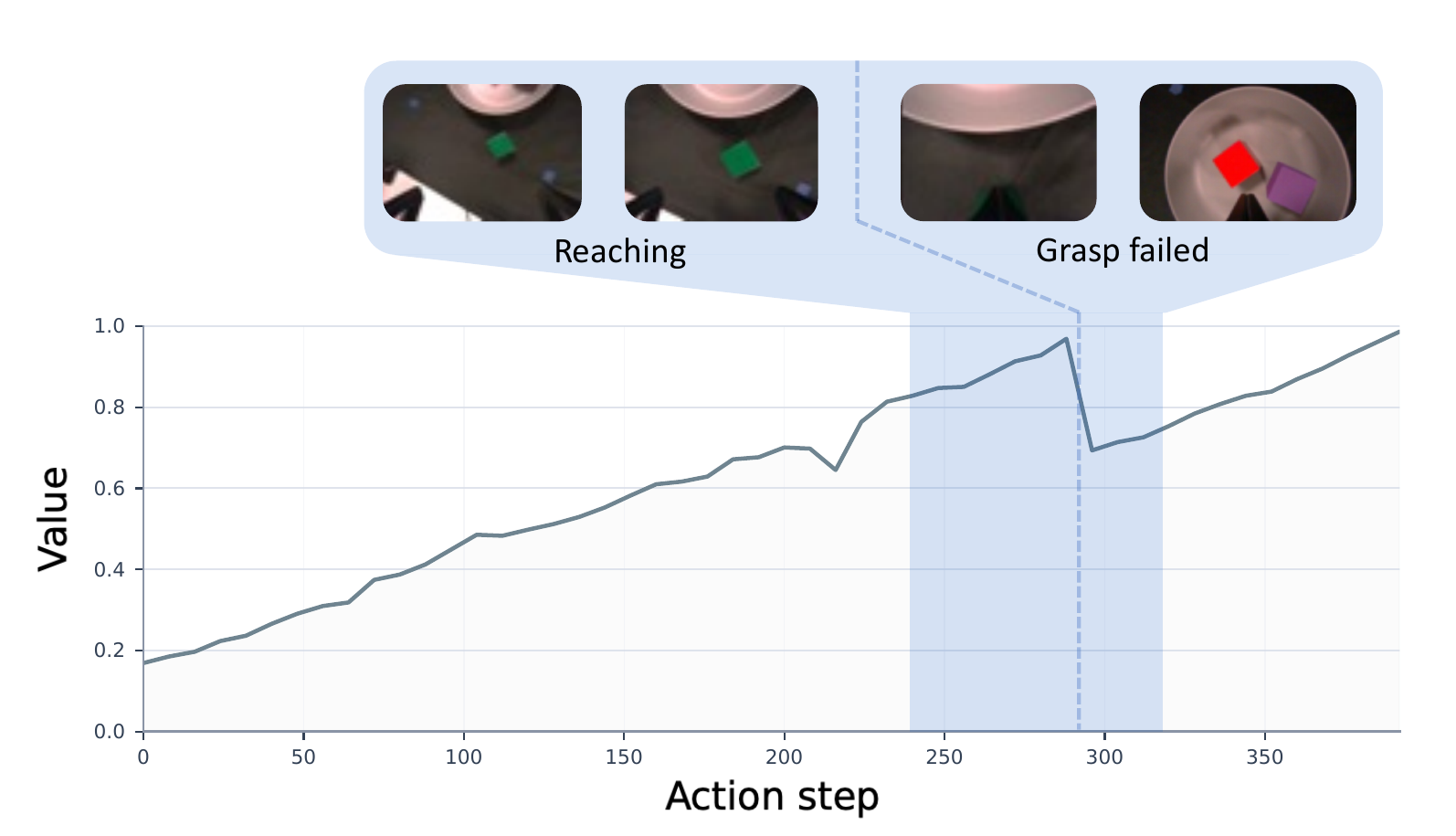}
        \vspace{-12pt}
        \caption{\textbf{Task-progress value over a rollout.} Value drops at a missed grasp and recovers after re-grasping, supporting its use as an action-conditioned ranking signal.}
        \label{fig:value_curve}
    \end{minipage}
    \vspace{-10pt}
\end{figure}

\paragraph{Value-guided candidate scoring.}
The action-conditioned task-progress value provides an optional ranking signal after it has been trained on failed outcomes. We sweep the number of candidates on a long-horizon grasping task using the same checkpoint, measuring both relative task completion and inference latency. As shown in Figure~\ref{fig:best_of_n}, task completion improves clearly from $N=1$ to $N=4$,
while latency increases with the number of candidates. Larger candidate sets provide a smaller gain relative to their additional cost, so optional scoring uses $N=4$ in our real-world experiments.

\paragraph{Value traces reflect action outcomes.}
Figure~\ref{fig:value_curve} visualizes a Pick Cubes rollout with a missed grasp followed by recovery. The predicted value rises as the task progresses, drops when the grasp fails, and increases again after the policy adjusts and re-grasps. Because the value is predicted after conditioning on the executed action, it can decrease at a later timestep when that action leads to a poor outcome.
Appendix~\ref{app:additional_value_traces} shows additional value traces, and Appendix~\ref{app:value_heat_map} further probes this behavior by scoring candidate placements in a controlled Stack Cubes grid.

%% file: sections/limit.tex

\section{Limitations}
\label{sec:limit}

While this work focuses on failure-aware consequence modeling, scaling the same causal training order to broader robot and human-interaction data may further improve the physical plausibility of future prediction and the discriminative power of the value head.
Also, future work could replace our value head or augment it with learned progress estimators while keeping the same action-conditioned value interface.
Finally, best-of-$N$ selection trades computation for reliability. \method{} can run in action-only mode when latency is critical, but value-guided selection requires an additional scoring pass for each candidate batch.

%% file: sections/conclusion.tex

\section{Conclusion}
\label{sec:conclusion}

We presented \method, a causal World-Action Model that reverses the usual WAM order by generating actions before predicting future video and task-progress value. A teacher-forcing mask makes the 
clean executed action the condition for all world-side predictions, 
allowing failure rollouts to supervise future and value prediction 
while their action imitation loss is disabled. 
Across simulation and real-world bimanual manipulation benchmarks, this failure-aware consequence modeling improves policy learning and reduces success-biased future hallucination under bad actions.
The learned progress predictor also provides an optional interface for candidate scoring, offering a compute-performance tradeoff at deployment.
This action-conditioned view of world modeling provides a natural interface for future training regimes that include online rollouts, DAgger-style corrections, and reinforcement learning from negative experience.

%% file: sections/appendix.tex

\clearpage
\appendix
\section*{Appendix}

\section{Training and Inference Algorithms}
\label{app:algorithms}

Algorithms~\ref{alg:cotrain}--\ref{alg:inference} summarize the procedure used by \method{}.
The notation follows Sec.~\ref{sec:method_training}: successful demonstrations are denoted by
$\mathcal{D}_s$, rollout failures by $\mathcal{D}_f$, and the action-conditioned task-progress
target is larger for actions that are expected to complete the task.

\begin{algorithm}[!htbp]
    \caption{Failure-aware co-training of \method{}}
    \label{alg:cotrain}
    \small
    \begin{algorithmic}[1]
        \Require successful demonstrations $\mathcal{D}_s$, failure rollouts $\mathcal{D}_f$, model $\theta$
        \For{each training step}
            \State Sample a minibatch $B_s \subset \mathcal{D}_s$ and $B_f \subset \mathcal{D}_f$
            \For{each trajectory window $(o_t,\ell,a_{t:t+H},o'_{t:t+K}) \in B_s \cup B_f$}
                \State Compute progress target $v_t(a_{t:t+H})$ using Eq.~\eqref{eq:failure_value}
                \State Pack tokens $[P,A,G,V,I]$ as in Eq.~\eqref{eq:sequence}
                \State Corrupt predicted action, value, and future-video targets with flow-matching noise
                \If{the window comes from $\mathcal{D}_f$}
                    \State Set action imitation mask $m_a \gets 0$
                \Else
                    \State Set action imitation mask $m_a \gets 1$
                \EndIf
            \EndFor
            \State Apply the teacher-forcing attention mask from Fig.~\ref{fig:arch_masks}
            \State Update $\theta$ with $m_a w_a\mathcal{L}_a + w_v\mathcal{L}_v + w_I\mathcal{L}_I$
        \EndFor
    \end{algorithmic}
\end{algorithm}

\begin{algorithm}[!htbp]
    \caption{Rollout failures for co-training}
    \label{alg:failure_collection}
    \small
    \begin{algorithmic}[1]
        \Require $\mathcal{D}_s$, initial policy $\pi_{\theta_0}$, rollout budget $M$
        \State Initialize $\mathcal{D}_f \gets \emptyset$
        \State Train $\pi_{\theta_0}$ on $\mathcal{D}_s$
        \For{task $\ell$ and rollout $m=1,\ldots,M$}
            \State $\tau_m=\{(o_t,a_t)\}_{t=1}^{T_m} \sim \pi_{\theta_0}(\cdot \mid o_t,\ell)$
            \If{$\mathrm{success}(\tau_m)=0$}
                \State Annotate failure onset $t_f$ when available
                \State $\mathcal{D}_f \gets \mathcal{D}_f \cup \{(\tau_m,\ell,t_f)\}$
            \EndIf
        \EndFor
        \State Continue training on $\mathcal{D}_s \cup \mathcal{D}_f$ using Algorithm~\ref{alg:cotrain}
    \end{algorithmic}
\end{algorithm}

\begin{algorithm}[!htbp]
    \caption{Two-stage inference with optional candidate scoring}
    \label{alg:inference}
    \small
    \begin{algorithmic}[1]
        \Require observation $o_t$, instruction $\ell$, number of candidates $N$
        \State Encode the observation prefix $P=(o_t,\ell)$
        \State Sample $N$ noisy action chunks $\{A^{(k)}\}_{k=1}^N$
        \For{$k=1,\ldots,N$ in parallel}
            \State \textbf{Stage 1:} denoise $A^{(k)}$ conditioned on $P$ to obtain $\hat{a}^{(k)}_{t:t+H}$
            \If{candidate scoring is disabled}
                \State \Return $\hat{a}^{(1)}_{t:t+H}$
            \EndIf
            \State \textbf{Stage 2:} place $\hat{a}^{(k)}_{t:t+H}$ in the clean action-conditioning slot
            \State Denoise the value token to obtain $\hat{v}^{(k)}=V_\theta(o_t,\ell,\hat{a}^{(k)}_{t:t+H})$
            \State Optionally denoise future-video tokens for consequence visualization
        \EndFor
        \State Select $k^\star = \arg\max_k \hat{v}^{(k)}$
        \State \Return $\hat{a}^{(k^\star)}_{t:t+H}$
    \end{algorithmic}
\end{algorithm}

\clearpage
\section{Task Overview}
\label{app:task_overview}

\begin{figure}[htbp]
    \centering
    \begin{minipage}{0.95\linewidth}
        \centering
        \includegraphics[width=\linewidth]{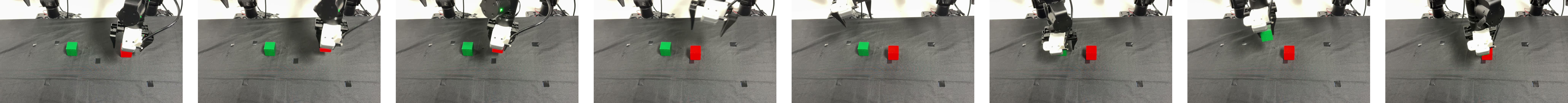}
        \vspace{0.5pt}

        \includegraphics[width=\linewidth]{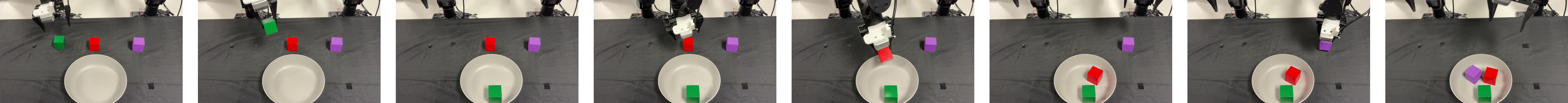}
        \vspace{0.5pt}

        \includegraphics[width=\linewidth]{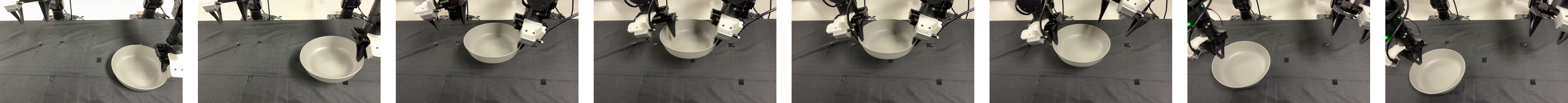}
        \vspace{0.5pt}

        \includegraphics[width=\linewidth]{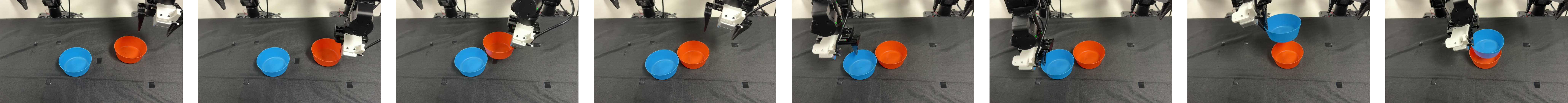}
        \vspace{0.5pt}

        \includegraphics[width=\linewidth]{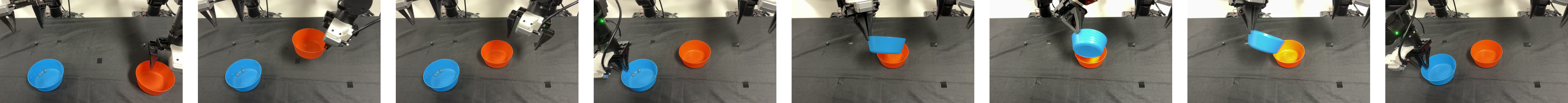}
    \end{minipage}
    \caption{\textbf{Seen real-world tasks.} Rollout image sequences for the five seen tasks reported in Table~\ref{tab:real_robot_seen}, ordered from top to bottom as Stack Cubes, Pick Cubes, Hand-over, Stack Bowls, and Pour.}
    \label{fig:task_overview_seen}
\end{figure}

\begin{figure}[htbp]
    \centering
    \begin{minipage}{0.95\linewidth}
        \centering
        \includegraphics[width=\linewidth]{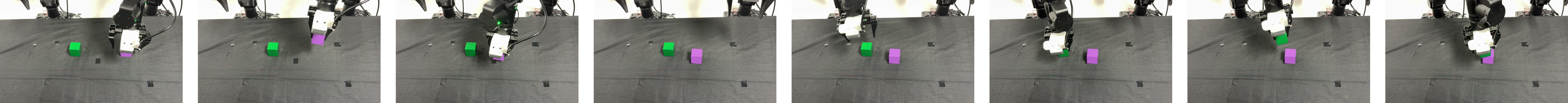}
        \vspace{0.5pt}

        \includegraphics[width=\linewidth]{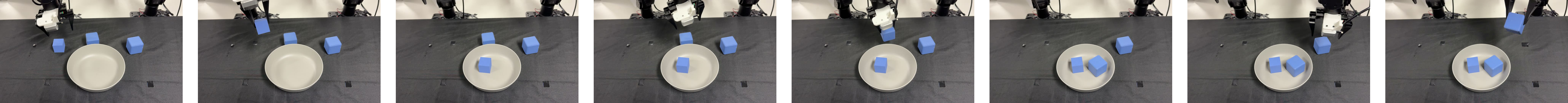}
        \vspace{0.5pt}

        \includegraphics[width=\linewidth]{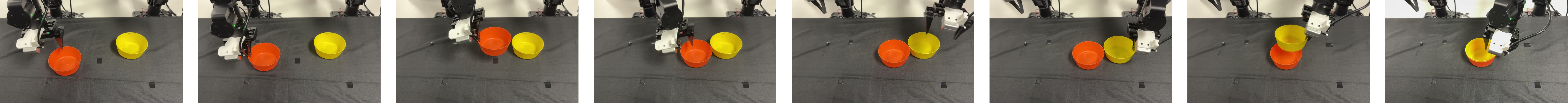}
    \end{minipage}
    \caption{\textbf{Unseen real-world tasks.} Held-out task variants used for Table~\ref{tab:real_robot_unseen}, ordered from top to bottom as Stack Cubes, Pick Cubes, and Stack Bowls.}
    \label{fig:task_overview_unseen}
\end{figure}

Figure~\ref{fig:task_overview_seen} and Figure~\ref{fig:task_overview_unseen}
show the real-world tasks evaluated in Sec.~\ref{sec:exp_real}. The seen set contains
five manipulation tasks used for real-world training and evaluation, and the unseen set
contains held-out variants with changed object colors, shapes, and instructions.

\section{Real-World Setup and Task Prompts}
\label{app:real_setup}

\definecolor{promptgray}{gray}{0.94}

\newcommand{\promptitem}[2]{\textbf{#1}\\#2\par\vspace{4pt}}

\begin{figure}[htbp]
    \centering
    \includegraphics[width=0.82\linewidth]{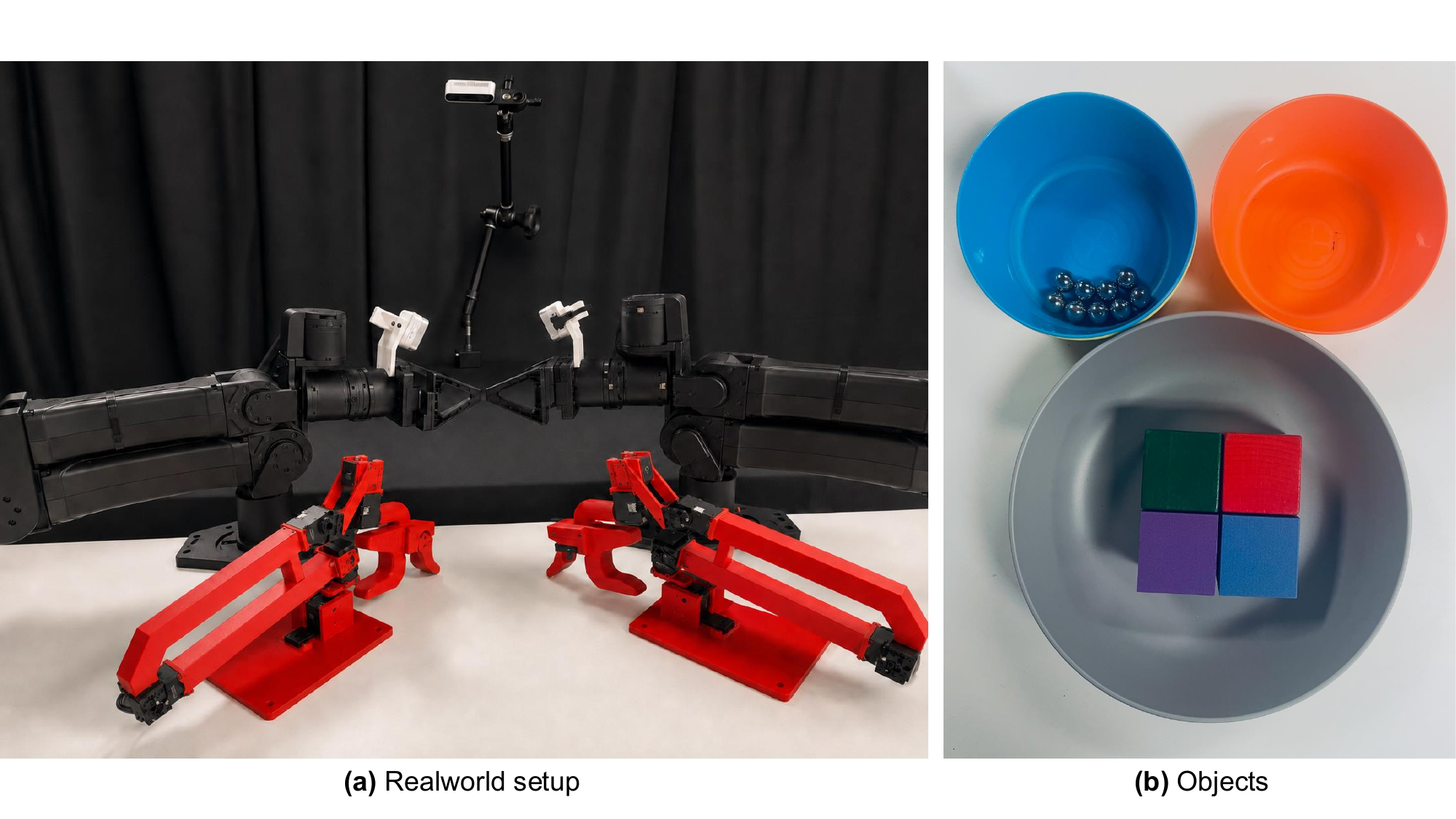}
    \caption{\textbf{Real-world hardware setup.} As shown in (a), we use two YAM arms, three RGB camera views, and a GELLO teleoperation device. The objects we used in real-world experiments are shown in (b).}
    \label{fig:hardware_setup}
\end{figure}

Figure~\ref{fig:hardware_setup} shows the real-world platform used for the experiments in
Sec.~\ref{sec:exp_real}. The setup contains two YAM robot arms, one Intel RealSense D435
camera for the main view, two D405 wrist cameras, and a red GELLO~\citep{wu2024gello} teleoperation device.
For multi-view observations, we pack the three camera streams into a single video canvas, preserving the policy observation while keeping the input compatible with the shared video backbone. 
We also pre-compute the VAE latent for each camera view to accelerate training.
The right side of the figure shows the manipulation objects used in the real-world tasks.

All baselines in Tables~\ref{tab:real_robot_seen} and~\ref{tab:real_robot_unseen} are fine-tuned on the same expert success demonstrations as \method{} before evaluation.
Table~\ref{tab:task_prompts} lists the language prompts used for the five seen real-world tasks.
These prompts correspond to the task names in Table~\ref{tab:real_robot_seen}.

\begin{table}[htbp]
    \centering
    \small
    \begingroup
    \setlength{\fboxsep}{0pt}
    \fcolorbox{black}{promptgray}{%
        \begin{minipage}{0.92\linewidth}
            \colorbox{black}{\makebox[\linewidth][l]{\hspace{8pt}\strut\textcolor{white}{\bfseries Real-world Task Prompts}}}
            \vspace{6pt}

            \hspace{8pt}\begin{minipage}{\dimexpr\linewidth-16pt\relax}
                \promptitem{Stack Cubes}{Grasp the red cube and place it at the center of the table, then grasp the green cube and stack it on top of the red cube.}
                \promptitem{Pick Cubes}{Pick up the red cube, green cube, and purple cube on the table and put them on the gray plate one by one.}
                \promptitem{Hand-over}{Use the left arm to pick up the gray plate on the table, and hand it over to the right arm and place it on the table.}
                \promptitem{Stack Bowls}{Grasp the orange bowl and place it at the center of the table, and then grasp the blue bowl and stack it on the orange bowl.}
                \promptitem{Pour}{Pick the orange bowl and put it at the center of the table, then pick the blue bowl and pour the little silver balls into the orange bowl.}
            \end{minipage}
            \vspace{5pt}
        \end{minipage}%
    }
    \endgroup
    \vspace{10pt}
    \caption{\textbf{Real-world task prompts.} For real-world experiments, we use the prompts listed above and compute the T5 embeddings for cross-attention.}
    \label{tab:task_prompts}
\end{table}

\clearpage
\section{Detailed RoboTwin Results}
\label{app:robotwin}

For RoboTwin training, we use a mixture of clean and domain-randomized demonstrations across the 50 tasks. The clean split contains 2,500 demonstrations in total, 
with 50 demonstrations per task, while the randomized split contains 25,000 demonstrations, with 500 demonstrations per task. The randomized scenes vary visual backgrounds, lighting, etc., providing a robustness test under distribution shift.

\begin{table}[p]
    \input{sections/robotwin_detailed_table}
    \caption{\textbf{Detailed RoboTwin results.} Per-task success rates on 50 RoboTwin tasks under clean and randomized evaluation. Each task is evaluated for 100 trials in each split.}
    \label{tab:robotwin_detailed_results}
\end{table}

\section{Success Rate and Inference Time}
\label{app:inference_time}

Table~\ref{tab:inference_time} compares success rates with measured deployment latency on an RTX PRO 6000.
Although Motus has the highest average success in simulation, its video-first inference is substantially slower.
\method{} keeps the action-first path lightweight: the action-only deployment used by the main policy runs faster
than recent WAM baselines while retaining strong simulation and real-world success.

\begin{table}[htbp]
    \centering
    \small
    \setlength{\tabcolsep}{8pt}
    \renewcommand{\arraystretch}{1.15}
    \caption{\textbf{Success rate and inference time.} Success rates are averages from Sec.~\ref{sec:exp}; latency is measured on an RTX PRO 6000.}
    \label{tab:inference_time}
    \begin{tabular}{@{}lccc@{}}
        \toprule
        \textbf{Model (inference time)} & \textbf{Sim} & \textbf{Real seen} & \textbf{Real unseen} \\
        \midrule
        $\pi_{0.5}$ (47 ms) & 79.8 & \underline{88} & \textbf{85} \\
        $\pi_0$ (45 ms) & 62.2 & 48 & 57 \\
        Motus (1220 ms) & \textbf{87.8} & 64 & 62 \\
        Cosmos (620 ms) & -- & 25 & 8 \\
        Ours w/ failure (380 ms) & \underline{87.5} & \textbf{89} & \underline{77} \\
        \bottomrule
    \end{tabular}
\end{table}

\section{Future Prediction Metrics}
\label{app:future_metrics}

Table~\ref{tab:future_prediction_full} reports the full future-image prediction metrics for the ablation in Sec.~\ref{sec:exp_ablation}, including both SSIM and PSNR. We evaluate on 512 held-out samples, split evenly between successful-demonstration windows and failure-rollout windows, and both models use 20 denoising steps. Consistent with the PSNR results reported in Table~\ref{tab:future_prediction_metrics}, failure-aware co-training substantially improves prediction quality on failure-rollout futures under both metrics while leaving successful-demonstration futures nearly unchanged.

\begin{table}[htbp]
    \centering
    \small
    \setlength{\tabcolsep}{8pt}
    \renewcommand{\arraystretch}{1.15}
    \caption{\textbf{Future prediction quality.} SSIM and PSNR on 512 held-out future-prediction samples, split evenly between successful-demonstration and failure-rollout windows. Both models use 20 denoising steps.}
    \label{tab:future_prediction_full}
    \begin{tabular}{@{}lcccc@{}}
    \toprule
    \textbf{Subset} & \multicolumn{2}{c}{\textbf{SSIM} $\uparrow$} & \multicolumn{2}{c}{\textbf{PSNR} $\uparrow$} \\
    \cmidrule(lr){2-3}\cmidrule(lr){4-5}
    & \textbf{Ours} & \textbf{Ours w/ failure} & \textbf{Ours} & \textbf{Ours w/ failure} \\
    \midrule
    All & 0.7873 & \textbf{0.8288} & 22.82 & \textbf{26.00} \\
    Success-rollout & 0.8285 & \textbf{0.8286} & \textbf{26.12} & 26.08 \\
    Failure-rollout & \cellcolor{red!12}0.7461 & \textbf{0.8290} & \cellcolor{red!12}19.51 & \textbf{25.92} \\
    \bottomrule
    \end{tabular}
\end{table}

\section{Additional Action-Conditioned Value Traces}
\label{app:additional_value_traces}

\begin{figure}[htbp]
    \centering
    \begin{minipage}[t]{0.48\linewidth}
        \centering
        \includegraphics[width=\linewidth]{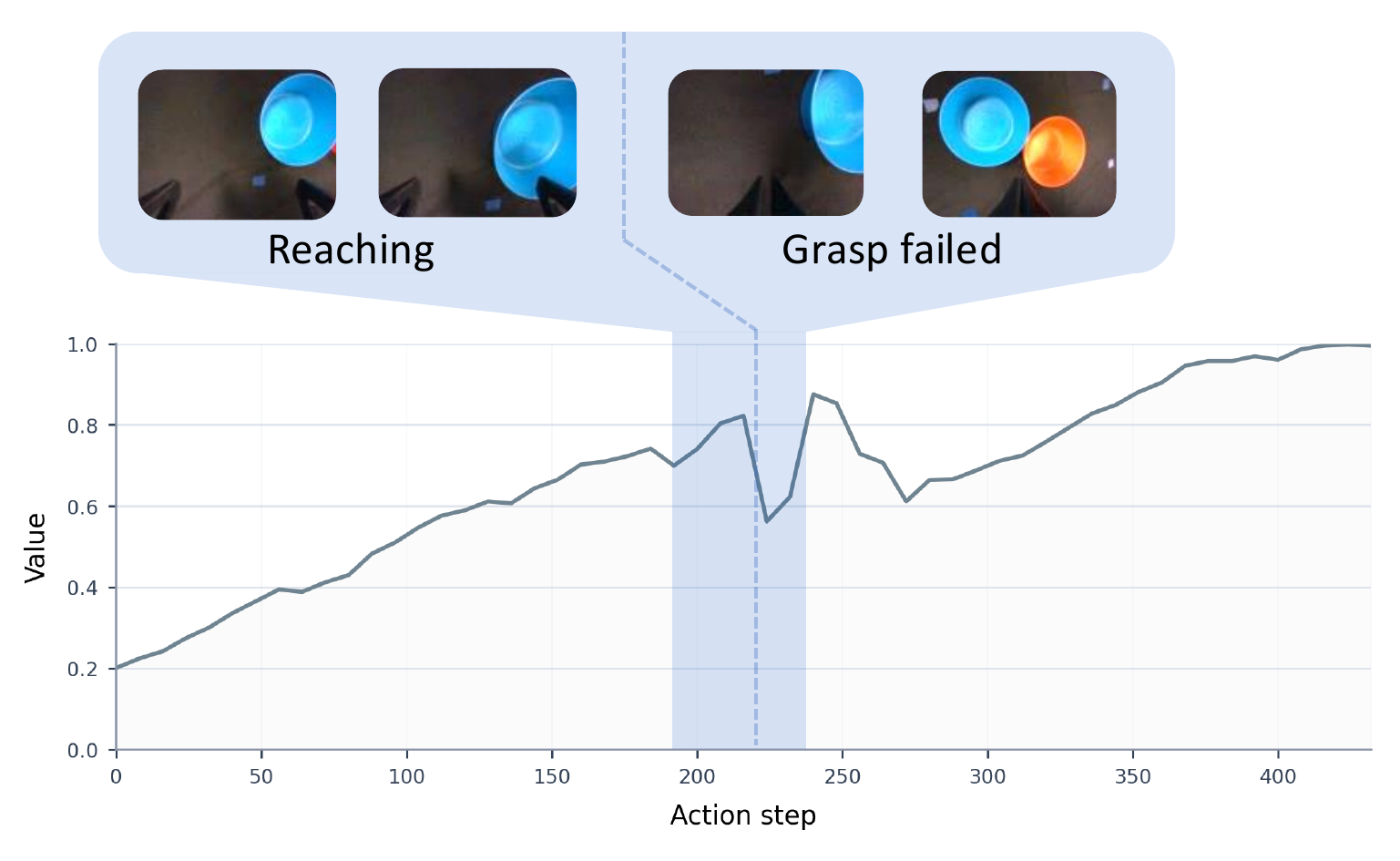}
    \end{minipage}
    \hfill
    \begin{minipage}[t]{0.48\linewidth}
        \centering
        \includegraphics[width=\linewidth]{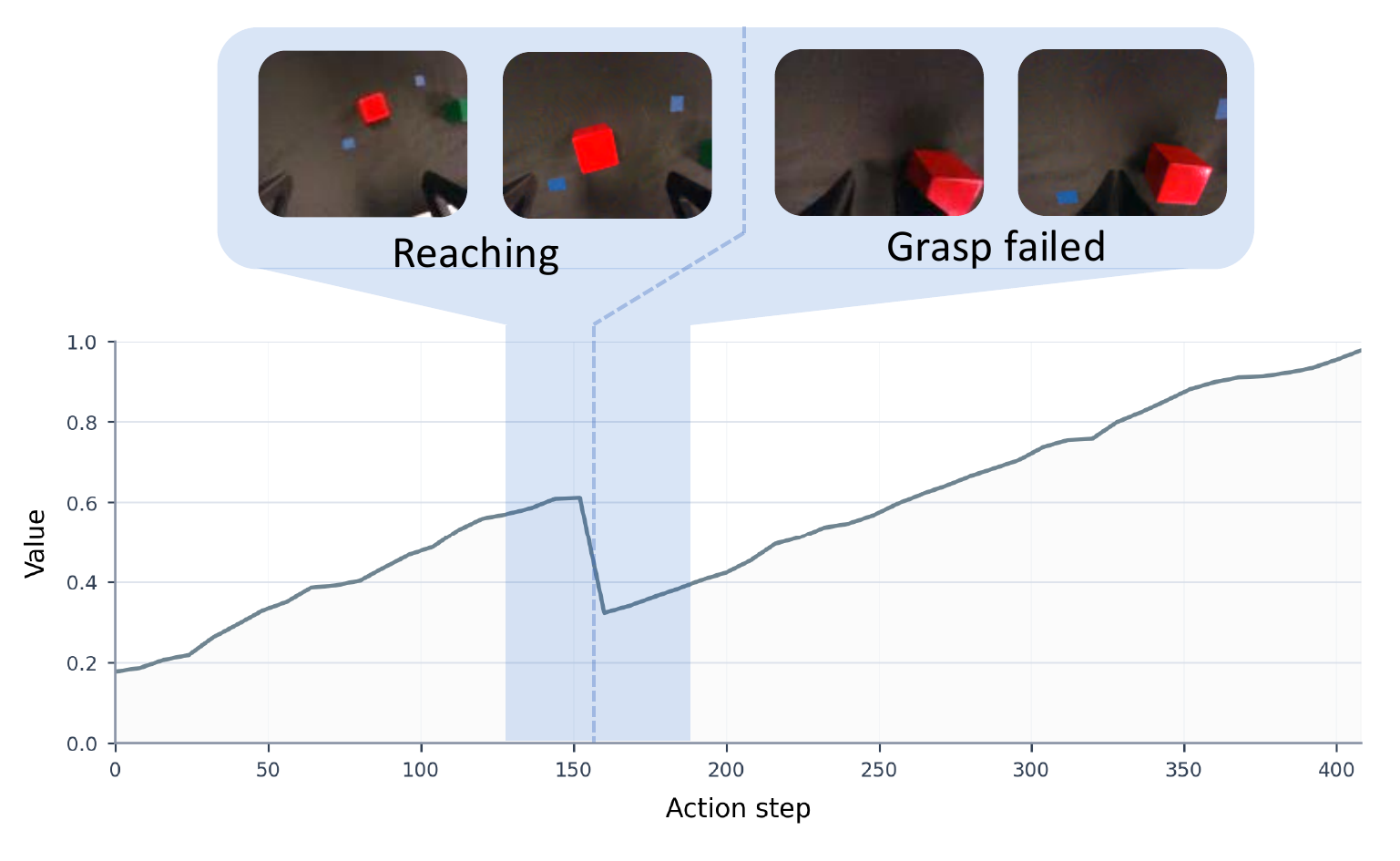}
    \end{minipage}
    \caption{\textbf{Additional action-conditioned value traces.} Predicted task-progress values over rollouts on Stack Bowls (left) and Stack Cubes (right).}
    \label{fig:additional_value_traces}
\end{figure}

Figure~\ref{fig:additional_value_traces} provides additional value-trace visualizations for the Stack Bowls and Stack Cubes tasks. Together with Figure~\ref{fig:value_curve}, these rollouts show that the predicted task-progress value changes with the action-conditioned task outcome across different manipulation skills.

\section{Action-Conditioned Value Heat Map}
\label{app:value_heat_map}

\begin{figure}[htbp]
    \centering
    \includegraphics[width=0.7\linewidth]{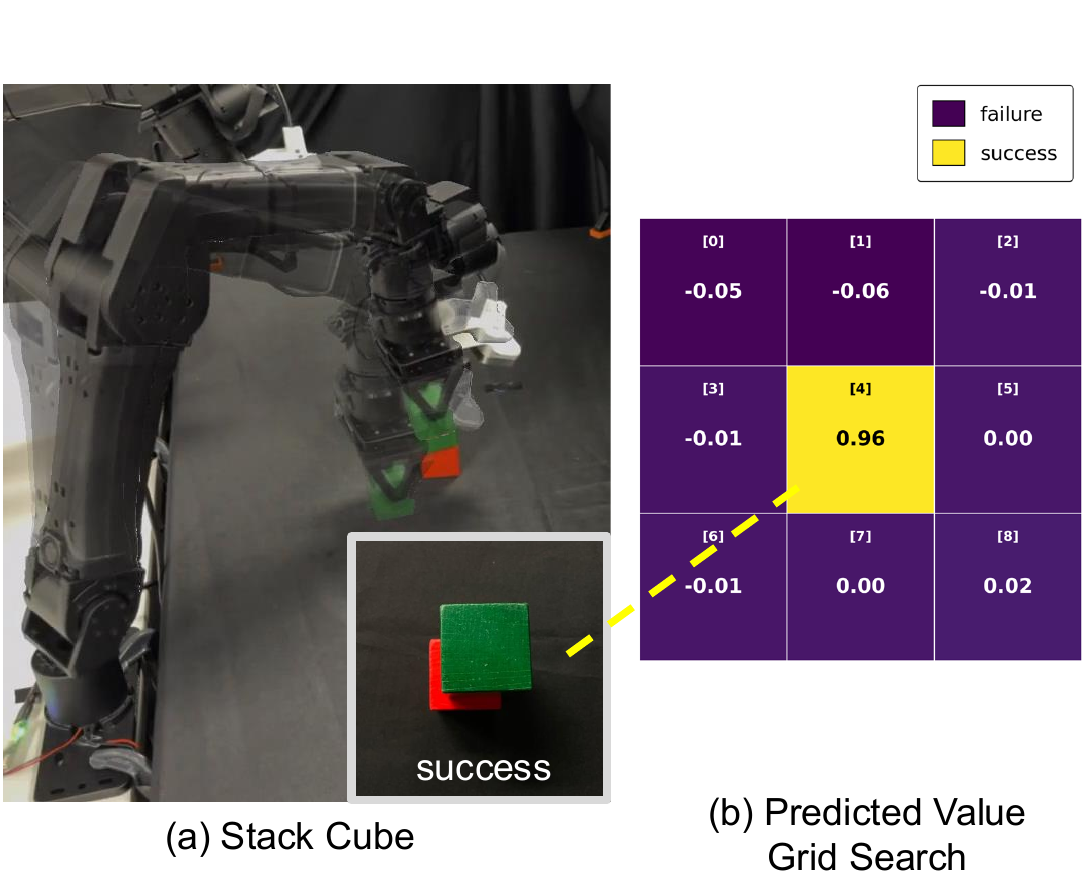}
    \caption{\textbf{Action-conditioned value heat map.} On Stack Cubes, the learned value head scores the successful center placement higher than failed placements around it.}
    \label{fig:value_heat_map}
\end{figure}

Figure~\ref{fig:value_heat_map} provides an additional diagnostic for the value head discussed in
Sec.~\ref{sec:exp_ablation}. On the Stack Cubes task, we evaluate the same state with candidate
placements over a $3\times3$ grid. Only the center placement completes the task successfully, shown in (a). The predicted
task-progress values are unclipped in this visualization, so failed action candidates can receive negative
scores. The model assigns the highest value to the successful center placement and lower values to placements
that move the cube to failure positions, supporting the action-conditioned target in Eq.~\eqref{eq:failure_value}.

%% file: sections/robotwin_detailed_table.tex

\centering
\scriptsize
\setlength{\tabcolsep}{3pt}
\renewcommand{\arraystretch}{0.92}
\resizebox{\textwidth}{!}{%
\begin{tabular}{@{}lcccccccccc@{}}
\toprule
\textbf{Simulation Task} &
\multicolumn{2}{c}{\textbf{X-VLA}} &
\multicolumn{2}{c}{\textbf{Motus}} &
\multicolumn{2}{c}{\shortstack{\textbf{\method{} w/o video}\\\textbf{co-train}}} &
\multicolumn{2}{c}{\textbf{\method{}}} &
\multicolumn{2}{c}{\shortstack{\textbf{\method{} w/}\\\textbf{failure}}} \\
\cmidrule(lr){2-3}
\cmidrule(lr){4-5}
\cmidrule(lr){6-7}
\cmidrule(lr){8-9}
\cmidrule(l){10-11}
& \textbf{Clean} & \textbf{Rand.}
& \textbf{Clean} & \textbf{Rand.}
& \textbf{Clean} & \textbf{Rand.}
& \textbf{Clean} & \textbf{Rand.}
& \textbf{Clean} & \textbf{Rand.} \\
\midrule
\textit{Adjust Bottle} & \textbf{100} & \textbf{99} & 89 & 93 & 96 & 97 & 99 & 98 & 99 & 98 \\
\textit{Beat Block Hammer} & 92 & \textbf{88} & \textbf{95} & \textbf{88} & 84 & 67 & 75 & 80 & 81 & 80 \\
\textit{Blocks Ranking Rgb} & 83 & 83 & \textbf{99} & \textbf{97} & 92 & 90 & 96 & 92 & 94 & 93 \\
\textit{Blocks Ranking Size} & 67 & \textbf{74} & \textbf{75} & 63 & 52 & 51 & 60 & 50 & 64 & 57 \\
\textit{Click Alarmclock} & 99 & 99 & \textbf{100} & \textbf{100} & 85 & 87 & 89 & 92 & 79 & 89 \\
\textit{Click Bell} & \textbf{100} & \textbf{100} & \textbf{100} & \textbf{100} & 68 & 89 & 93 & 91 & 80 & 90 \\
\textit{Dump Bin Bigbin} & 79 & 77 & 95 & 91 & 93 & 91 & \textbf{97} & 95 & 96 & \textbf{100} \\
\textit{Grab Roller} & \textbf{100} & \textbf{100} & \textbf{100} & \textbf{100} & \textbf{100} & \textbf{100} & \textbf{100} & \textbf{100} & \textbf{100} & \textbf{100} \\
\textit{Handover Block} & 73 & 37 & \textbf{86} & \textbf{73} & 68 & 55 & 65 & 62 & 84 & 72 \\
\textit{Handover Mic} & 0 & 0 & 78 & 63 & 96 & 96 & 99 & \textbf{100} & \textbf{100} & 99 \\
\textit{Hanging Mug} & 23 & 27 & \textbf{38} & 38 & 35 & 38 & 36 & 45 & 36 & \textbf{49} \\
\textit{Lift Pot} & 99 & \textbf{100} & 96 & 99 & 97 & 99 & 99 & \textbf{100} & \textbf{100} & 99 \\
\textit{Move Can Pot} & \textbf{89} & 86 & 34 & 74 & 82 & 82 & 80 & 86 & 80 & \textbf{90} \\
\textit{Move Pillbottle Pad} & 73 & 71 & 93 & 96 & 94 & 91 & 94 & 91 & \textbf{97} & \textbf{97} \\
\textit{Move Playingcard Away} & 93 & 98 & \textbf{100} & 96 & \textbf{100} & 95 & \textbf{100} & 97 & \textbf{100} & \textbf{100} \\
\textit{Move Stapler Pad} & 78 & 73 & \textbf{83} & \textbf{85} & 40 & 41 & 54 & 50 & 62 & 50 \\
\textit{Open Laptop} & 93 & \textbf{100} & 95 & 91 & 97 & 99 & \textbf{98} & 97 & \textbf{98} & \textbf{100} \\
\textit{Open Microwave} & 79 & 71 & 95 & 91 & 87 & 78 & 91 & 95 & \textbf{98} & \textbf{98} \\
\textit{Pick Diverse Bottles} & 58 & 36 & \textbf{90} & \textbf{91} & 71 & 66 & 65 & 58 & 79 & 67 \\
\textit{Pick Dual Bottles} & 47 & 36 & 96 & \textbf{90} & 88 & 80 & 87 & 74 & \textbf{97} & 83 \\
\textit{Place A2b Left} & 48 & 49 & 88 & 79 & 89 & \textbf{89} & 92 & \textbf{89} & \textbf{95} & \textbf{89} \\
\textit{Place A2b Right} & 36 & 36 & 91 & 87 & 86 & 90 & 90 & 90 & \textbf{93} & \textbf{93} \\
\textit{Place Bread Basket} & 81 & 71 & \textbf{91} & \textbf{94} & 86 & 76 & 84 & 80 & 85 & 74 \\
\textit{Place Bread Skillet} & 77 & 67 & \textbf{86} & \textbf{83} & 85 & 79 & \textbf{86} & 81 & 84 & 81 \\
\textit{Place Burger Fries} & 94 & 94 & \textbf{98} & 98 & 95 & \textbf{99} & \textbf{98} & 97 & \textbf{98} & 95 \\
\textit{Place Can Basket} & 49 & 52 & 81 & \textbf{76} & 75 & 67 & 85 & 69 & \textbf{87} & 71 \\
\textit{Place Cans Plasticbox} & 97 & 98 & 98 & 94 & \textbf{99} & 95 & \textbf{99} & 97 & \textbf{99} & \textbf{100} \\
\textit{Place Container Plate} & 97 & 95 & 98 & \textbf{99} & 98 & 96 & 96 & 98 & \textbf{100} & 98 \\
\textit{Place Dual Shoes} & 79 & \textbf{88} & \textbf{93} & 87 & 62 & 65 & 83 & 85 & 85 & 80 \\
\textit{Place Empty Cup} & \textbf{100} & 98 & 99 & 98 & 97 & \textbf{100} & 99 & \textbf{100} & \textbf{100} & \textbf{100} \\
\textit{Place Fan} & 80 & 75 & 91 & 87 & 87 & 75 & 87 & \textbf{88} & \textbf{95} & 87 \\
\textit{Place Mouse Pad} & 70 & 70 & 66 & 68 & 52 & 65 & 69 & 77 & \textbf{86} & \textbf{82} \\
\textit{Place Object Basket} & 44 & 39 & 81 & \textbf{87} & 76 & 79 & \textbf{90} & 81 & 88 & 86 \\
\textit{Place Object Scale} & 52 & 74 & \textbf{88} & \textbf{85} & 83 & 74 & 81 & 80 & 84 & 84 \\
\textit{Place Object Stand} & 86 & 88 & \textbf{98} & \textbf{97} & 91 & 93 & 96 & 94 & \textbf{98} & 94 \\
\textit{Place Phone Stand} & 88 & 87 & 87 & 86 & 89 & 92 & 89 & \textbf{95} & \textbf{90} & 91 \\
\textit{Place Shoe} & 96 & 95 & \textbf{99} & 97 & 98 & 95 & \textbf{99} & \textbf{99} & 98 & 98 \\
\textit{Press Stapler} & 92 & \textbf{98} & \textbf{93} & \textbf{98} & 76 & 74 & 84 & 73 & 82 & 79 \\
\textit{Put Bottles Dustbin} & 74 & 77 & 81 & 79 & 63 & 75 & 73 & 81 & \textbf{83} & \textbf{88} \\
\textit{Put Object Cabinet} & 46 & 48 & 88 & 71 & 79 & 78 & \textbf{94} & \textbf{85} & 89 & 82 \\
\textit{Rotate Qrcode} & 34 & 33 & \textbf{89} & 73 & 79 & 79 & 79 & 82 & 81 & \textbf{84} \\
\textit{Scan Object} & 14 & 36 & 67 & 66 & 78 & 72 & \textbf{87} & 77 & 86 & \textbf{80} \\
\textit{Shake Bottle Horizontally} & \textbf{100} & \textbf{100} & \textbf{100} & 98 & \textbf{100} & 98 & \textbf{100} & 99 & \textbf{100} & 98 \\
\textit{Shake Bottle} & 99 & \textbf{100} & \textbf{100} & 97 & 99 & 95 & \textbf{100} & 99 & \textbf{100} & 97 \\
\textit{Stack Blocks Three} & 6 & 10 & 91 & \textbf{95} & 83 & 79 & 88 & 91 & \textbf{96} & 94 \\
\textit{Stack Blocks Two} & 92 & 87 & \textbf{100} & \textbf{98} & 97 & 96 & 95 & 97 & \textbf{100} & 95 \\
\textit{Stack Bowls Three} & 76 & 86 & 79 & \textbf{87} & 74 & 70 & \textbf{82} & 75 & 79 & 77 \\
\textit{Stack Bowls Two} & 96 & 93 & \textbf{98} & \textbf{98} & 97 & 95 & 97 & 93 & 94 & 93 \\
\textit{Stamp Seal} & 76 & 82 & \textbf{93} & \textbf{92} & 60 & 69 & 72 & 80 & 78 & 91 \\
\textit{Turn Switch} & 40 & 61 & \textbf{84} & \textbf{78} & 69 & 48 & 66 & 61 & 61 & 56 \\
\midrule
\textbf{\textit{Average}} & 72.88 & 72.84 & \textbf{88.66} & \textbf{87.02} & 82.54 & 80.98 & 86.34 & 84.92 & \underline{88.36} & \underline{86.56} \\
\bottomrule
\end{tabular}%
}

%% file: reference.bib
@article{hu2026bagelvla,
  title={Bagelvla: Enhancing long-horizon manipulation via interleaved vision-language-action generation},
  author={Hu, Yucheng and Zhang, Jianke and Luo, Yuanfei and Guo, Yanjiang and Chen, Xiaoyu and Sun, Xinshu and Feng, Kun and Lu, Qingzhou and Chen, Sheng and Zhang, Yangang and others},
  journal={arXiv preprint arXiv:2602.09849},
  year={2026}
}

@article{li2026lintbot,
  title={Causal World Modeling for Robot Control},
  author={Li, Lin and Zhang, Qihang and Luo, Yiming and Yang, Shuai and Wang, Ruilin and Han, Fei and Yu, Mingrui and Gao, Zelin and Xue, Nan and Zhu, Xing and others},
  journal={arXiv preprint arXiv:2601.21998},
  year={2026}
}

@article{yuan2026fast,
  title={Fast-WAM: Do World Action Models Need Test-time Future Imagination?},
  author={Yuan, Tianyuan and Dong, Zibin and Liu, Yicheng and Zhao, Hang},
  journal={arXiv preprint arXiv:2603.16666},
  year={2026}
}

@article{ye2026gigaworld,
  title={GigaWorld-Policy: An Efficient Action-Centered World--Action Model},
  author={Ye, Angen and Wang, Boyuan and Ni, Chaojun and Huang, Guan and Zhao, Guosheng and Li, Hao and Li, Hengtao and Li, Jie and Lv, Jindi and Liu, Jingyu and others},
  journal={arXiv preprint arXiv:2603.17240},
  year={2026}
}

@article{kim2026cosmos,
  title={Cosmos policy: Fine-tuning video models for visuomotor control and planning},
  author={Kim, Moo Jin and Gao, Yihuai and Lin, Tsung-Yi and Lin, Yen-Chen and Ge, Yunhao and Lam, Grace and Liang, Percy and Song, Shuran and Liu, Ming-Yu and Finn, Chelsea and others},
  journal={arXiv preprint arXiv:2601.16163},
  year={2026}
}

@article{bi2025motus,
  title={Motus: A unified latent action world model},
  author={Bi, Hongzhe and Tan, Hengkai and Xie, Shenghao and Wang, Zeyuan and Huang, Shuhe and Liu, Haitian and Zhao, Ruowen and Feng, Yao and Xiang, Chendong and Rong, Yinze and others},
  journal={arXiv preprint arXiv:2512.13030},
  year={2025}
}

@article{lv2026viva,
  title={ViVa: A Video-Generative Value Model for Robot Reinforcement Learning},
  author={Lv, Jindi and Li, Hao and Li, Jie and Nie, Yifei and Kong, Fankun and Wang, Yang and Wang, Xiaofeng and Zhu, Zheng and Ni, Chaojun and Deng, Qiuping and others},
  journal={arXiv preprint arXiv:2604.08168},
  year={2026}
}

@article{black2024pi0,
  title={$\pi_0$: A Vision-Language-Action Flow Model for General Robot Control},
  author={Black, Kevin and Brown, Noah and Driess, Danny and Esmail, Adnan and Equi, Michael and Finn, Chelsea and Fusai, Niccolo and Groom, Lachy and Hausman, Karol and Ichter, Brian and others},
  journal={arXiv preprint arXiv:2410.24164},
  year={2024}
}

@article{physicalintelligence2025pi05,
  title={$\pi_{0.5}$: A Vision-Language-Action Model with Open-World Generalization},
  author={{Physical Intelligence} and Black, Kevin and Brown, Noah and Darpinian, James and Dhabalia, Karan and Driess, Danny and Esmail, Adnan and Equi, Michael and Finn, Chelsea and Fusai, Niccolo and others},
  journal={arXiv preprint arXiv:2504.16054},
  year={2025}
}

@article{chen2025robotwin2,
  title={RoboTwin 2.0: A Scalable Data Generator and Benchmark with Strong Domain Randomization for Robust Bimanual Robotic Manipulation},
  author={Chen, Tianxing and Chen, Zanxin and Chen, Baijun and Cai, Zijian and Liu, Yibin and Li, Zixuan and Liang, Qiwei and Lin, Xianliang and Ge, Yiheng and Gu, Zhenyu and others},
  journal={arXiv preprint arXiv:2506.18088},
  year={2025}
}

@article{zheng2025xvla,
  title={X-vla: Soft-prompted transformer as scalable cross-embodiment vision-language-action model},
  author={Zheng, Jinliang and Li, Jianxiong and Wang, Zhihao and Liu, Dongxiu and Kang, Xirui and Feng, Yuchun and Zheng, Yinan and Zou, Jiayin and Chen, Yilun and Zeng, Jia and others},
  journal={arXiv preprint arXiv:2510.10274},
  year={2025}
}

@inproceedings{brohan2023rt2,
  title={Rt-2: Vision-language-action models transfer web knowledge to robotic control},
  author={Zitkovich, Brianna and Yu, Tianhe and Xu, Sichun and Xu, Peng and Xiao, Ted and Xia, Fei and Wu, Jialin and Wohlhart, Paul and Welker, Stefan and Wahid, Ayzaan and others},
  booktitle={Conference on Robot Learning},
  pages={2165--2183},
  year={2023},
  organization={PMLR}
}

@article{kim2024openvla,
  title={Openvla: An open-source vision-language-action model},
  author={Kim, Moo Jin and Pertsch, Karl and Karamcheti, Siddharth and Xiao, Ted and Balakrishna, Ashwin and Nair, Suraj and Rafailov, Rafael and Foster, Ethan and Lam, Grace and Sanketi, Pannag and others},
  journal={arXiv preprint arXiv:2406.09246},
  year={2024}
}

@article{wan2025wan,
  title={Wan: Open and advanced large-scale video generative models},
  author={Wan, Team and Wang, Ang and Ai, Baole and Wen, Bin and Mao, Chaojie and Xie, Chen-Wei and Chen, Di and Yu, Feiwu and Zhao, Haiming and Yang, Jianxiao and others},
  journal={arXiv preprint arXiv:2503.20314},
  year={2025}
}

@article{loshchilov2017decoupled,
  title={Decoupled weight decay regularization},
  author={Loshchilov, Ilya and Hutter, Frank},
  journal={arXiv preprint arXiv:1711.05101},
  year={2017}
}

@article{ye2026world,
  title={World action models are zero-shot policies},
  author={Ye, Seonghyeon and Ge, Yunhao and Zheng, Kaiyuan and Gao, Shenyuan and Yu, Sihyun and Kurian, George and Indupuru, Suneel and Tan, You Liang and Zhu, Chuning and Xiang, Jiannan and others},
  journal={arXiv preprint arXiv:2602.15922},
  year={2026}
}

@article{brohan2022rt,
  title={Rt-1: Robotics transformer for real-world control at scale},
  author={Brohan, Anthony and Brown, Noah and Carbajal, Justice and Chebotar, Yevgen and Dabis, Joseph and Finn, Chelsea and Gopalakrishnan, Keerthana and Hausman, Karol and Herzog, Alex and Hsu, Jasmine and others},
  journal={arXiv preprint arXiv:2212.06817},
  year={2022}
}

@article{team2024octo,
  title={Octo: An open-source generalist robot policy},
  author={Team, Octo Model and Ghosh, Dibya and Walke, Homer and Pertsch, Karl and Black, Kevin and Mees, Oier and Dasari, Sudeep and Hejna, Joey and Kreiman, Tobias and Xu, Charles and others},
  journal={arXiv preprint arXiv:2405.12213},
  year={2024}
}

@article{wu2026pragmatic,
  title={A Pragmatic VLA Foundation Model},
  author={Wu, Wei and Lu, Fan and Wang, Yunnan and Yang, Shuai and Liu, Shi and Wang, Fangjing and Zhu, Qian and Sun, He and Wang, Yong and Ma, Shuailei and others},
  journal={arXiv preprint arXiv:2601.18692},
  year={2026}
}

@article{lingbotvla2,
      title={From Foundation to Application: Improving VLA Models in Practice}, 
      author={Wei Wu and Fangjing Wang and Fan Lu and He Sun and Shi Liu and Yunnan Wang and Yibin Yan and Yong Wang and Shuailei Ma and Xinyang Wang and Yibin Liu and Shuai Yang and Tianxiang Zhou and Kejia Zhang and Lei Zhou and Cheng Su and Nan Xue and Bin Tan and Han Zhang and Youchao Zhang and Fei Liao and Xing Zhu and Yujun Shen and Kecheng Zheng},
      journal={arXiv preprint arXiv:2607.06403},
      year={2026}
}

@inproceedings{o2024open,
  title={Open x-embodiment: Robotic learning datasets and rt-x models},
  author={O’Neill, Abby and Rehman, Abdul and Maddukuri, Abhiram and Gupta, Abhishek and Padalkar, Abhishek and Lee, Abraham and Pooley, Acorn and Gupta, Agrim and Mandlekar, Ajay and Jain, Ajinkya and others},
  booktitle={2024 IEEE International Conference on Robotics and Automation (ICRA)},
  pages={6892--6903},
  year={2024},
  organization={IEEE}
}

@article{gao2026dreamdojo,
  title={DreamDojo: A Generalist Robot World Model from Large-Scale Human Videos},
  author={Gao, Shenyuan and Liang, William and Zheng, Kaiyuan and Malik, Ayaan and Ye, Seonghyeon and Yu, Sihyun and Tseng, Wei-Cheng and Dong, Yuzhu and Mo, Kaichun and Lin, Chen-Hsuan and others},
  journal={arXiv preprint arXiv:2602.06949},
  year={2026}
}

@article{yang2026rise,
  title={Rise: Self-improving robot policy with compositional world model},
  author={Yang, Jiazhi and Lin, Kunyang and Li, Jinwei and Zhang, Wencong and Lin, Tianwei and Wu, Longyan and Su, Zhizhong and Zhao, Hao and Zhang, Ya-Qin and Chen, Li and others},
  journal={arXiv preprint arXiv:2602.11075},
  year={2026}
}

@article{dasari2019robonet,
  title={Robonet: Large-scale multi-robot learning},
  author={Dasari, Sudeep and Ebert, Frederik and Tian, Stephen and Nair, Suraj and Bucher, Bernadette and Schmeckpeper, Karl and Singh, Siddharth and Levine, Sergey and Finn, Chelsea},
  journal={arXiv preprint arXiv:1910.11215},
  year={2019}
}

@inproceedings{walke2023bridgedata,
  title={Bridgedata v2: A dataset for robot learning at scale},
  author={Walke, Homer Rich and Black, Kevin and Zhao, Tony Z and Vuong, Quan and Zheng, Chongyi and Hansen-Estruch, Philippe and He, Andre Wang and Myers, Vivek and Kim, Moo Jin and Du, Max and others},
  booktitle={Conference on Robot Learning},
  pages={1723--1736},
  year={2023},
  organization={PMLR}
}

@article{khazatsky2024droid,
  title={Droid: A large-scale in-the-wild robot manipulation dataset},
  author={Khazatsky, Alexander and Pertsch, Karl and Nair, Suraj and Balakrishna, Ashwin and Dasari, Sudeep and Karamcheti, Siddharth and Nasiriany, Soroush and Srirama, Mohan Kumar and Chen, Lawrence Yunliang and Ellis, Kirsty and others},
  journal={arXiv preprint arXiv:2403.12945},
  year={2024}
}

@article{zhao2023learning,
  title={Learning fine-grained bimanual manipulation with low-cost hardware},
  author={Zhao, Tony Z and Kumar, Vikash and Levine, Sergey and Finn, Chelsea},
  journal={arXiv preprint arXiv:2304.13705},
  year={2023}
}

@article{fu2024mobile,
  title={Mobile aloha: Learning bimanual mobile manipulation with low-cost whole-body teleoperation},
  author={Fu, Zipeng and Zhao, Tony Z and Finn, Chelsea},
  journal={arXiv preprint arXiv:2401.02117},
  year={2024}
}

@article{chi2024universal,
  title={Universal manipulation interface: In-the-wild robot teaching without in-the-wild robots},
  author={Chi, Cheng and Xu, Zhenjia and Pan, Chuer and Cousineau, Eric and Burchfiel, Benjamin and Feng, Siyuan and Tedrake, Russ and Song, Shuran},
  journal={arXiv preprint arXiv:2402.10329},
  year={2024}
}

@article{xu2025dexumi,
  title={Dexumi: Using human hand as the universal manipulation interface for dexterous manipulation},
  author={Xu, Mengda and Zhang, Han and Hou, Yifan and Xu, Zhenjia and Fan, Linxi and Veloso, Manuela and Song, Shuran},
  journal={arXiv preprint arXiv:2505.21864},
  year={2025}
}

@article{cheng2026tacumi,
  title={TacUMI: A Multi-Modal Universal Manipulation Interface for Contact-Rich Tasks},
  author={Cheng, Tailai and Chen, Kejia and Chen, Lingyun and Zhang, Liding and Zhang, Yue and Ling, Yao and Hamad, Mahdi and Bing, Zhenshan and Wu, Fan and Sharma, Karan and others},
  journal={arXiv preprint arXiv:2601.14550},
  year={2026}
}

@article{yang2025egovla,
  title={Egovla: Learning vision-language-action models from egocentric human videos},
  author={Yang, Ruihan and Yu, Qinxi and Wu, Yecheng and Yan, Rui and Li, Borui and Cheng, An-Chieh and Zou, Xueyan and Fang, Yunhao and Cheng, Xuxin and Qiu, Ri-Zhao and others},
  journal={arXiv preprint arXiv:2507.12440},
  year={2025}
}

@article{zheng2026egoscale,
  title={Egoscale: Scaling dexterous manipulation with diverse egocentric human data},
  author={Zheng, Ruijie and Niu, Dantong and Xie, Yuqi and Wang, Jing and Xu, Mengda and Jiang, Yunfan and Casta{\~n}eda, Fernando and Hu, Fengyuan and Tan, You Liang and Fu, Letian and others},
  journal={arXiv preprint arXiv:2602.16710},
  year={2026}
}

@article{nasiriany2024robocasa,
  title={Robocasa: Large-scale simulation of everyday tasks for generalist robots},
  author={Nasiriany, Soroush and Maddukuri, Abhiram and Zhang, Lance and Parikh, Adeet and Lo, Aaron and Joshi, Abhishek and Mandlekar, Ajay and Zhu, Yuke},
  journal={arXiv preprint arXiv:2406.02523},
  year={2024}
}

@article{nasiriany2026robocasa365,
  title={Robocasa365: A large-scale simulation framework for training and benchmarking generalist robots},
  author={Nasiriany, Soroush and Nasiriany, Sepehr and Maddukuri, Abhiram and Zhu, Yuke},
  journal={arXiv preprint arXiv:2603.04356},
  year={2026}
}

@article{mandlekar2023mimicgen,
  title={Mimicgen: A data generation system for scalable robot learning using human demonstrations},
  author={Mandlekar, Ajay and Nasiriany, Soroush and Wen, Bowen and Akinola, Iretiayo and Narang, Yashraj and Fan, Linxi and Zhu, Yuke and Fox, Dieter},
  journal={arXiv preprint arXiv:2310.17596},
  year={2023}
}

@article{liu2023libero,
  title={Libero: Benchmarking knowledge transfer for lifelong robot learning},
  author={Liu, Bo and Zhu, Yifeng and Gao, Chongkai and Feng, Yihao and Liu, Qiang and Zhu, Yuke and Stone, Peter},
  journal={Advances in Neural Information Processing Systems},
  volume={36},
  pages={44776--44791},
  year={2023}
}

@article{mees2022calvin,
  title={Calvin: A benchmark for language-conditioned policy learning for long-horizon robot manipulation tasks},
  author={Mees, Oier and Hermann, Lukas and Rosete-Beas, Erick and Burgard, Wolfram},
  journal={IEEE Robotics and Automation Letters},
  volume={7},
  number={3},
  pages={7327--7334},
  year={2022},
  publisher={IEEE}
}

@article{li2025vla,
  title={Vla-rft: Vision-language-action reinforcement fine-tuning with verified rewards in world simulators},
  author={Li, Hengtao and Ding, Pengxiang and Suo, Runze and Wang, Yihao and Ge, Zirui and Zang, Dongyuan and Yu, Kexian and Sun, Mingyang and Zhang, Hongyin and Wang, Donglin and others},
  journal={arXiv preprint arXiv:2510.00406},
  year={2025}
}

@inproceedings{wu2024gello,
  title={Gello: A general, low-cost, and intuitive teleoperation framework for robot manipulators},
  author={Wu, Philipp and Shentu, Yide and Yi, Zhongke and Lin, Xingyu and Abbeel, Pieter},
  booktitle={2024 IEEE/RSJ International Conference on Intelligent Robots and Systems (IROS)},
  pages={12156--12163},
  year={2024},
  organization={IEEE}
}

@article{bai2024hallucination,
  title={Hallucination of multimodal large language models: A survey},
  author={Bai, Zechen and Wang, Pichao and Xiao, Tianjun and He, Tong and Han, Zongbo and Zhang, Zheng and Shou, Mike Zheng},
  journal={arXiv preprint arXiv:2404.18930},
  year={2024}
}

@article{rawte2023survey,
  title={A survey of hallucination in large foundation models},
  author={Rawte, Vipula and Sheth, Amit and Das, Amitava},
  journal={arXiv preprint arXiv:2309.05922},
  year={2023}
}

@article{lyu2026lda,
  title={Lda-1b: Scaling latent dynamics action model via universal embodied data ingestion},
  author={Lyu, Jiangran and Liu, Kai and Zhang, Xuheng and Liao, Haoran and Feng, Yusen and Zhu, Wenxuan and Shen, Tingrui and Chen, Jiayi and Zhang, Jiazhao and Dong, Yifei and others},
  journal={arXiv preprint arXiv:2602.12215},
  year={2026}
}

@article{jiang2026cross,
  title={Cross-Hand Latent Representation for Vision-Language-Action Models},
  author={Jiang, Guangqi and Liang, Yutong and Ye, Jianglong and Huang, Jia-Yang and Jing, Changwei and Duan, Rocky and Abbeel, Pieter and Wang, Xiaolong and Zou, Xueyan},
  journal={arXiv preprint arXiv:2603.10158},
  year={2026}
}

@article{liu2026long,
  title={Long-Horizon Manipulation via Trace-Conditioned VLA Planning},
  author={Liu, Isabella and Cheng, An-Chieh and Yan, Rui and Chen, Geng and Qiu, Ri-Zhao and Zou, Xueyan and Yi, Sha and Yin, Hongxu and Wang, Xiaolong and Liu, Sifei},
  journal={arXiv preprint arXiv:2604.21924},
  year={2026}
}

@article{qiu2025humanoid,
  title={Humanoid policy\~{} human policy},
  author={Qiu, Ri-Zhao and Yang, Shiqi and Cheng, Xuxin and Chawla, Chaitanya and Li, Jialong and He, Tairan and Yan, Ge and Yoon, David J and Hoque, Ryan and Paulsen, Lars and others},
  journal={arXiv preprint arXiv:2503.13441},
  year={2025}
}

@article{qi2026inference,
  title={Inference-Time Enhancement of Generative Robot Policies via Predictive World Modeling},
  author={Qi, Han and Yin, Haocheng and Zhu, Aris and Du, Yilun and Yang, Heng},
  journal={IEEE Robotics and Automation Letters},
  year={2026},
  publisher={IEEE}
}

@article{lipman2022flow,
  title={Flow matching for generative modeling},
  author={Lipman, Yaron and Chen, Ricky TQ and Ben-Hamu, Heli and Nickel, Maximilian and Le, Matt},
  journal={arXiv preprint arXiv:2210.02747},
  year={2022}
}

@article{zhou2026tau,
  title={$\tau_0$-WM: A Unified Video-Action World Model for Robotic Manipulation},
  author={Zhou, Pengfei and Chen, Shengcong and Chen, Di and Wang, Jiaxu and Jin, Rongjun and Zhu, Bingwen and Pan, Yike and Gu, Songen and Wang, Kuanning and Nan, Shufeng and others},
  journal={arXiv preprint arXiv:2606.01027},
  year={2026}
}

@article{liu2023reflect,
  title={Reflect: Summarizing robot experiences for failure explanation and correction},
  author={Liu, Zeyi and Bahety, Arpit and Song, Shuran},
  journal={arXiv preprint arXiv:2306.15724},
  year={2023}
}

@article{grollman2012robot,
  title={Robot learning from failed demonstrations},
  author={Grollman, Daniel H and Billard, Aude G},
  journal={International Journal of Social Robotics},
  volume={4},
  number={4},
  pages={331--342},
  year={2012},
  publisher={Springer}
}

@article{wang2026learning,
  title={Learning from Demonstration with Failure Awareness for Safe Robot Navigation},
  author={Wang, Xianghui and Cheng, Siwei and Wang, Shanze and Zhang, Xinming and Zhang, Dan and Zhang, Wei},
  journal={arXiv preprint arXiv:2604.23360},
  year={2026}
}

@article{li2026recover,
  title={Recover, Discover, Plan: Learning Skills and Concepts from Robot Failures},
  author={Li, Bowen and Mishra, Mayank and Liu, Y Isabel and Tao, Stone and Kumar, Nishanth and Gray, Alexander G and Wickramarachchi, Ruwan and Francis, Jonathan and Scherer, Sebastian and Silver, Tom},
  journal={arXiv preprint arXiv:2606.18328},
  year={2026}
}
